\documentclass{article}

\usepackage[preprint]{neurips_2025}

\usepackage{microtype}
\usepackage{graphicx}
\usepackage{subfigure}
\usepackage{booktabs} 
\usepackage{dor_macros_icml}
\usepackage{wrapfig} 

\usepackage{subcaption} 

\usepackage{algorithmicx}
\usepackage{algpseudocode}   
\usepackage{caption}

\usepackage{hyperref}

\usepackage{algorithm}

\usepackage{amsmath}
\usepackage{amssymb}
\usepackage{mathtools}
\usepackage{amsthm}

\usepackage[capitalize,noabbrev]{cleveref}

\theoremstyle{plain}
\newtheorem{theorem}{Theorem}[section]
\newtheorem{proposition}[theorem]{Proposition}
\newtheorem{lemma}[theorem]{Lemma}

\theoremstyle{definition}

\theoremstyle{remark}
\newtheorem{remark}[theorem]{Remark}

\usepackage[textsize=tiny]{todonotes}

\usepackage[utf8]{inputenc} 
\usepackage[T1]{fontenc}    
\usepackage{hyperref}       
\usepackage{url}            
\usepackage{booktabs}       
\usepackage{amsfonts}       
\usepackage{nicefrac}       
\usepackage{microtype}      
\usepackage{xcolor}         

\usepackage[numbers]{natbib}

\title{Neural Estimation for Scaling\\Entropic Multimarginal Optimal Transport}

\author{%
  Dor Tsur \\
  Ben-Gurion University \\
  \texttt{dortz@post.bgu.ac.il} 
  \And
  Ziv Goldfeld \\
  Cornell University\\
  \texttt{goldfeld@cornell.edu} 
  \And
  Kristjan Greenewald \\
  MIT-IBM Watson AI Lab\\
  \texttt{kristjan.h.greenewald@ibm.com} 
  \And
  Haim Permuter \\
  Ben-Gurion University\\
  \texttt{haimp@bgu.ac.il} 
}

\begin{document}

\maketitle

\begin{abstract}
Multimarginal optimal transport (MOT) is a powerful framework for modeling interactions between multiple distributions, yet its applicability is bottlenecked by a high computational overhead. Entropic regularization provides computational speedups via the multimarginal Sinkhorn algorithm, whose time complexity, for a dataset size $n$ and $k$ marginals, generally scales as $O(n^k)$. However, this dependence on the dataset size $n$ is computationally prohibitive for many machine learning problems. In this work, we propose a new computational framework for entropic MOT, dubbed Neural Entropic MOT (NEMOT), that enjoys significantly improved scalability. NEMOT employs neural networks trained using mini-batches, which transfers the computational complexity from the dataset size to the size of the mini-batch, leading to substantial gains. We provide formal guarantees on the accuracy of NEMOT via non-asymptotic error bounds. We supplement these with numerical results that demonstrate the performance gains of NEMOT over Sinkhorn's algorithm, as well as extensions to neural computation of multimarginal entropic Gromov-Wasserstein alignment. In particular, orders-of-magnitude speedups are observed relative to the state-of-the-art, with a notable increase in the feasible number of samples and marginals. NEMOT seamlessly integrates as a module in large-scale machine learning pipelines, and can serve to expand the practical applicability of entropic MOT for tasks involving multimarginal~data.
\end{abstract}

\section{Introduction}
Optimal transport (OT) \cite{villani2009optimal,santambrogio2015optimal} is a versatile framework for modeling complex relationships between distributions, which has had a tangible impact in machine learning \cite{solomon2015convolutional, tolstikhin2017wasserstein, arjovsky2017wasserstein}, statistics \cite{carlier2016vector,chernozhukov2017monge}, economics \cite{galichon2018optimal}, and applied mathematics \cite{santambrogio2015optimal}. In many applications, one seeks to account for interaction between more than two marginals, which naturally leads to the multimarginal OT (MOT) setting \cite{pass2015multi}. Given $k \geq 2$ marginal distributions $\muk\coloneqq(\mu_1, \dots, \mu_k)$ each supported on $\cX_1,\ldots,\cX_k$, respectively, and cost function $c:\cX_1\times\cdots\times\cX_k\to[0,\infty)$, the MOT problem reads
\begin{equation}\label{eq:MOT}
    \MOT_c(\muk) \coloneqq \inf_{\pi \in \Pi(\muk)} \int c(\mathbf{x}^k) d\pi(\mathbf{x}^k),
\end{equation}
where $\Pi(\muk)$ is the set of all joint distributions on $\bigotimes_{i=1}^k\cX_i$ whose $i$th marginal is $\mu_i$, for~$i=1,\ldots,k$. 

MOT was initially employed for certain operational problems in physics \cite{cotar2013density,benamou2019generalized}, but more recently it has found applications in machine learning, spanning multi-domain image translation \cite{cao2019multi}, Bayesian learning \cite{srivastava2018scalable}, distributionally robust optimization \cite{chen2022distributionally}, adversarial learning \cite{trillos2023multimarginal}, fair representation learning \cite{mehta2023efficient} and multiview embedding matching \cite{piran2024contrasting}. However, the scalability of such algorithms has been hindered by the computational overhead of MOT, which requires solving an exponentially large (in $k$) linear program. In fact, \cite{altschuler2021hardness} show that a polynomial-time deterministic algorithm for MOT does not exist for a variety of cost structures, even when the cost tensor is sparse. 



In the bimarginal case, entropic regularization is key for the empirical successes of OT. By leveraging Sinkhorn's fixed-point iteration algorithm \cite{cuturi2013sinkhorn}, entropic OT reduces the computational complexity while enjoying fast statistical estimation rates \cite{genevay2019sample,mena2019statistical}. Entropic regularization also leads to computational gains in the multimarginal setting via an extension of Sinkhorn's algorithm \cite{lin2022complexity}. However, multimarginal Sinkhorn directly computes a $k$-wise joint distribution (as an $n^k$-sized tensor), resulting in an exponential in $k$ scaling of the memory and computational costs. One approach to reduce this complexity is to restrict attention to sparse cost structures, where the $k$-fold cost function in \eqref{eq:MOT} decomposes into a sum of a small number of pairwise costs (e.g., circle or tree cost structures). Such sparsity reduces the runtime of Sinkhorn to polynomial in $(n, k)$ \cite{haasler2021multi,ba2022accelerating,altschuler2023polynomial}, but the dependence on the entire dataset size $n$ remains. This is computationally prohibitive for large real-world datasets.

\subsection{Contributions}
We develop a methodology for scaling up the entropic MOT (EMOT) computation. Drawing upon recent neural estimation approaches \cite{belghazi2018mutual,sreekumar2022neural,tsur2023neural,wang2024neural}, we propose the neural EMOT (NEMOT) framework, which optimizes an approximation of the dual EMOT objective over a set of neural network parameters. NEMOT estimates both the EMOT cost and transportation plan by harnessing deep learning tools, such as mini-batch gradient-based optimization. This transfers the computational cost of our algorithm to the mini-batch size $b$ (as opposed to $n$), which results in a marked increase in scalability compared to multimarginal Sinkhorn. Furthermore, while existing approaches often rely on restrictive assumptions on the cost function (e.g., sparsity) to alleviate some computational burden, NEMOT can handle massive datasets with arbitrary costs. As such, NEMOT can be seamlessly integrated as a module into larger machine learning pipelines as a loss or~regularizer.



To facilitate a principled implementation of NEMOT, we provide non-asymptotic bounds on the cost and plan estimation erros. Under mild assumptions, a NEMOT realized by a $m$-neuron shallow ReLU network and $n$ data samples from each of the $k$ distributions is guaranteed to estimate the cost $\MOT_c(\muk)$ within an additive error of $O(\mathsf{poly}(\epsilon^{-1})k(m^{-\frac{1}{2}}+n^{-\frac{1}{2}}))$. We further show, that the same bound, up to an extra multiplicative factor of $1/\epsilon$, holds for the Kullback-Leibler (KL) divergence between the optimal EMOT plan and its neural estimate. The theory is supplemented by empirical results that demonstrate the computational advantage of NEMOT over the multimarginal Sinkhorn algorithm. NEMOT exhibits speedups in orders of magnitude, unlocking new regimes of possible $(b,n,k)$ values for EMOT and applications thereof. We present extensions of NEMOT to the multimarginal Gromov-Wasserstein (MGW) problem, which gained recent popularity for tasks involving alignment of heterogeneous datasets \cite{alvarez2018gromov,xu2019gromov}. To that end, we reformulate the MGW as an alternating optimization between NEMOT and certain auxiliary matrices. 

\subsection{Related Work} 
MOT has seen increasing applications as a means to alter the distribution of learned representations in a way that enforces criteria across multiple domains. In generative modeling, \cite{cao2019multi} introduces a multimarginal extension of the Wasserstein GAN \cite{arjovsky2017wasserstein} to address the multimarginal matching problem. In multi-view learning, the multimarginal Sinkhorn algorithm is employed for distribution matching of multiple learned embeddings \cite{piran2024contrasting}. \cite{mehta2023efficient} leveraged MOT with $1$-Wasserstein cost to regularize data representations, with applications to fairness, harmonization, and generative modeling. In the context of adversarial classification, \cite{trillos2023multimarginal} shows the equivalence between MOT and certain adversarial multiclass classification problems, allowing the derivation of optimal robust classification rules. More recently, \cite{noble2024tree} proposed to solve high-dimensional star-structured EMOT problems via diffusion processes. However, this approach does not generalize to other cost functions, leaving barycenter computation as its main application. For unregularized MOT, \cite{zhou2024efficient} suggests solving a simplified surrogate problem over a directed tree, but their computational complexity also scales with~$n$.

When only two marginals are present, EMOT reduces to the ubiquitous entropic optimal transport problem (EOT). Neural estimation of EOT has seen significant success via several different approaches. For instance, \cite{seguy2017large} introduced a neural estimation framework based on the semi-dual formulation of the entropic OT problem. Their method efficiently computes the optimal transport plan by parametrizing the dual potentials with neural networks. Building on this, \cite{daniels2021score} proposed a score‐based generative approach that not only estimates the Sinkhorn coupling via neural network parametrization but also incorporates stochastic dynamics to sample from the resulting transport plan. More recently, \cite{mokrov2023energy} combined the semi-dual entropic OT formulation with energy-based models, further increasing modeling capacity by capturing complex cost structures through expressive neural networks. Theoretical aspects of neural entropic OT were studied in \cite{wang2024neural}, providing sharp estimation bounds on the optimal transport cost and approximation guarantees for the neural bimarginal transport plan. We anticipate that similar advances will be attainable in the multimarginal setting, building on the neural estimation approach proposed herein.


Overall, existing EMOT estimation methods for more than two marginals (i.e., excluding the reduction to EOT) suffer from two main problems: (1) most approaches are tailored to specific cost functions, which restricts applicability \cite{beier2023multi,noble2024tree}, and (2) they often suffer from scalability issues, with computational complexity growing as $O(n^k)$ \cite{piran2024contrasting,zhou2024efficient}. The NEMOT framework we propose addresses these issues, facilitating broad applicability of MOT methods at scales not possible before. 

\section{Background}
\paragraph{Entropic multimarginal optimal transport.}
Consider a set of $k$ marginals $\muk$, each $\mu_i$ is supported on $\cX_i$, for $i=1,\ldots,k$, and define $\cxk\coloneqq\cX_1\times\cdots\times\cX_k$. 
We assume that $\cX_i\subset\RR^{d_i}$ is compact for all $i=1,\ldots,k$.
For a cost function $c:\cxk\to\RR$, the EMOT is defined as 
\begin{equation}\label{eq:EMOT}
    \MOT_{c,\epsilon}(\muk)\coloneqq\mspace{-7mu}\inf_{\pi\in\Pi(\muk)}\int \mspace{-3mu}c(\bxk)d\pi(\bxk) + \epsilon\DKL(\pi\|\otimes_{i=1}^k\mu_i),
\end{equation}
where $\mathbf{x}^k=(x_1,\ldots,x_k)\in\cxk$, $\DKL(\mu\|\nu)\coloneqq\EE_\mu[\log(d\mu/d\nu)]$ is the KL divergence between probability measures $\mu\ll\nu$ (defined as $+\infty$ in the absence of absolute continuity), and $\Pi(\muk)$ is the set of joint distributions over $\cxk$ with marginal $\mu_i$ for every $i=1,\ldots,k$. EMOT is a convexification of the linear MOT program \eqref{eq:MOT}, which attains a unique optimal plan $\pi_\epsilon$ that can be computed via a multimarginal Sinkhorn algorithm \cite{lin2022complexity}. As such, EMOT improves upon the complexity of unregularized MOT, with the cost approximation gap generally scaling as $O(\epsilon\log(1/\epsilon))$ \cite{nenna2023convergence}.

\begin{figure}[!b]
    \centering
    \includegraphics[trim={100pt 110pt 110pt 100pt},clip,width=0.5\linewidth]{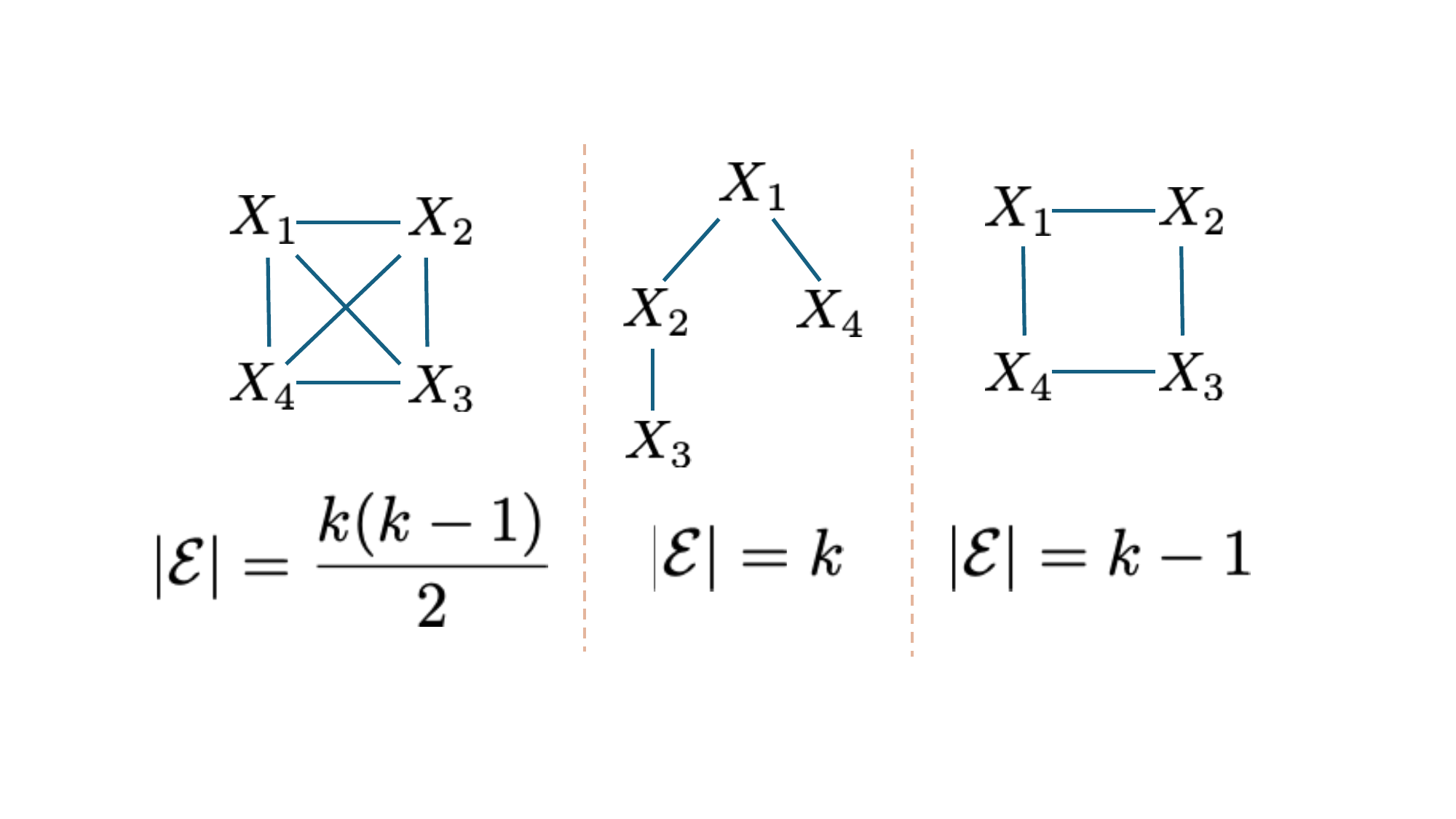}
    \caption{Cost graphs with $|\cE|$ edges.}
    \label{fig:cost_graph}
\end{figure}

EMOT attains the following dual representation \cite{carlier2020differential}, which is key for the proposed NEMOT methods:
\begin{align}\label{eq:nmot_dual}
    \MOT_{c,\epsilon}(\muk)=\sup_{\varphi_1,\dots,\varphi_k}\sum_{i=1}^k\int_{\cX_i}\varphi_i d\mu_i
    -\epsilon\int_{\cxk} e^{\left(\bigoplus_{i=1}^k\varphi_i -c\right)/\epsilon}d(\otimes_{i=1}^k\mu_i)+\epsilon,
\end{align}
where $\bigoplus_{i=1}^k\varphi_i(\bxk)\coloneqq\sum_{i=1}^k\varphi_i(x_i)$ is the direct sum of the dual potentials $\varphi_i:\cX_i\to\RR$, $i=1,\dots, k$. Notably, \eqref{eq:nmot_dual} is an unconstrained concave maximization and the optimal $k$-tuple $(\varphi_1^\star,\ldots,\varphi_k^\star)$ is unique up to additive constants. The unique EMOT plan can be represented in its terms as 
\begin{equation}\label{eq:opt_plan}
d\pi_\epsilon = \exp\left(\frac{\bigoplus_{i=1}^k\varphi_i^\star - c}{\epsilon}\right)d(\otimes_{i=1}^k \mu_i).
\end{equation}
EMOT is often solved using the multimarginal Sinkhorn algorithm \cite{lin2022complexity}, which generalizes the matrix scaling procedure \cite{sinkhorn1967diagonal} to tensor data structures. However, the computational bottleneck of Sinkhorn stems from a marginalization step of the coupling tensor, whose complexity is generally $O(n^k)$. This is a key limitation in the applicability of EMOT to large datasets that this paper aims to overcome.


  

\paragraph{Graphical representation of the cost.}
Consider a class of MOT cost functions that decompose into a sum of pairwise terms,\footnote{For simplicity of presentation we use the same pairwise cost function $\tilde{c}$ for all pairs, although the framework readily extends to edge-dependent pairwise cost functions.} i.e., $c(\mathbf{x}^k)=\sum_{i,j}\tilde{c}(x_i,x_j)$. This enables representing the cost function using a graph $\cG_c=(\cV,\cE)$, where $\cV=\{1,\ldots,k\}$ and the edge set $\cE$ corresponds to the pairwise components in the decomposition.
Namely, $(i,j)\in\cE$ if the term $\tilde{c}(x_i,x_j)$ appears in the cost decomposition. The size of $\cE$ is induced by the structure of $c(\bxk)$ and determines the complexity of the coupling marginalization. In the general case, where $\cE$ contains all pairs $i\neq j$, $|\cE|$ grows as $k^2$ and the corresponding Sinkhorn iteration complexity remains $O(n^k)$. We refer to such $c(\bxk)$ as a \textit{full cost}. 
See Figure \ref{fig:cost_graph} for cost graph examples.
Imposing structure on the graph can reduce the scaling with $k$, as the coupling marginalization can be computed via matrix-matrix operations \cite{haasler2021multi,ba2022accelerating,altschuler2023polynomial}. For instance, for circular or tree graphs, $|\cE|= O(k)$ and 
the Sinkhorn time complexity becomes polynomial in $(n,k)$.

\begin{figure}[!t]
  \centering
  \begin{minipage}[t]{0.45\textwidth}
    \vspace{0pt}  
    \captionsetup{type=algorithm}
    \hrule height 0.4pt
    \captionof{algorithm}{NEMOT}\label{alg:nemot}
    \hrule height 0.3pt
    \begin{algorithmic}[1]
      \State \textbf{Input:} dataset $\mathbf{X}^{n,k}$
      \State \textbf{Output:} estimated EMOT, neural plan
      \State initialise $\bm{\theta}=(\theta_i)_{i=1}^k$
      \Repeat
        \State sample batch $\mathbf{X}^{b,k}$
        \State compute
          $\widehat{\MOT}_{c,\epsilon}(\mathbf{X}^{b,k})$
          via~\eqref{eq:ne_emot}
        \State $\bm{\theta}\gets\bm{\theta}+
               \nabla_{\bm{\theta}}
               \widehat{\MOT}_{c,\epsilon}(\mathbf{X}^{b,k})$
      \Until{convergence}\\
      \Return $\widehat{\MOT}_{c,\epsilon}(\mathbf{X}^{n,k})$,
              plan $\pi_{\epsilon}^{\bm{\theta}^\star}$
    \end{algorithmic}
    \hrule height 0.4pt
  \end{minipage}\hfill
  \begin{minipage}[t]{0.55\textwidth}
    \vspace{3mm}
    \centering
    \hspace{2mm}\includegraphics[trim=80pt 90pt 50pt 15pt,clip,width=0.93\linewidth]
      {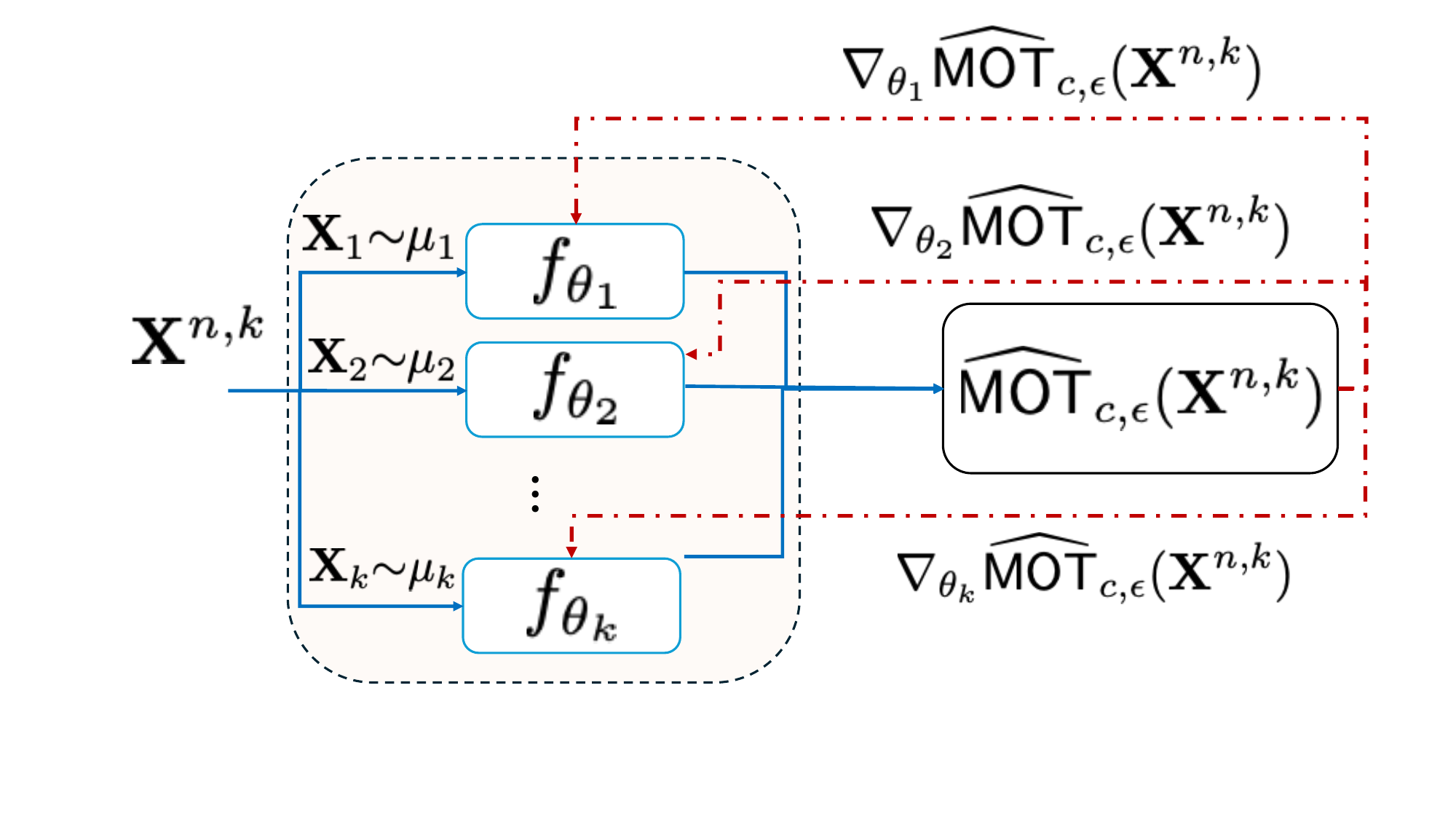}
  \end{minipage}

  \caption{%
    NEMOT algorithm and system architecture. The diagram shows how each marginal dataset $\mathbf{X}_j$, $j=1,\dots,k$, is fed to its corresponding neural potential $f_{\theta_j}$; the outputs are aggregated into the NEMOT loss \eqref{eq:ne_emot}, from which gradients are calculated for parameter~update.}
  \label{fig:nemot_combined}
\end{figure}

\section{Neural Estimation of Entropic MOT}
The NEMOT algorithm enables scaling up EMOT computation to large and high-dimensional~problems. The idea is to parameterize the set of dual potentials in \eqref{eq:nmot_dual} by feedforward neural nets, approximate expectations using sample means, and optimize the resulting objective over parameter space. Let $\mathbf{X}^{n,k}=\{\mathbf{X}^k_1,\ldots,\mathbf{X}^k_n\}$ be the full dataset, where each $\mathbf{X}^k_i=(X_{i1}\ldots X_{ik})$ comprises $n$ independent and identically distributed (i.i.d.) $k$-tuples sampled from $\mu_1\otimes\cdots\otimes\mu_k$. 
NEMOT~takes~the~form
\begin{align}\label{eq:ne_emot}
    &\widehat{\MOT}_{c,\epsilon}(\textbf{X}^{n,k}) \coloneqq \sup_{\theta_1,\dots,\theta_k}\frac{1}{n}\sum_{i=1}^n \sum_{j=1}^k f_{\theta_j}(X_{ij})-\frac{\epsilon}{n}\sum_{i=1}^n\exp\left(\frac{\sum_{j=1}^k f_{\theta_j}(X_{ij}) -c(\mathbf{X}^k_i)}{\epsilon}\right)+\epsilon,
\end{align}
where $f_{\theta_i}:\cX_i\to\RR$ is a neural net with parameters $\theta_i\in\RR^{d'_i}$. We optimize the above objective using mini-batch gradient descent over the parameters $\bm{\theta}\coloneqq(\theta_1,\dots,\theta_k)$. The NEMOT pipeline is depicted in Figure \ref{fig:nemot_combined} and the training procedure is summarized in Algorithm \ref{alg:nemot}.

Once converged, the learned parameters $\bm{\theta}^\star$ specify the $k$-tuple of neural dual potentials $f_{\theta_1^\star},\dots,f_{\theta_k^\star}$.\footnote{We may reduce to $k-1$ neural nets by representing one network as the $(c,\epsilon)$-transform~of~the~rest.} The EMOT estimate is obtained by evaluating the objective in \eqref{eq:ne_emot} on the entire~dataset $\textbf{X}^{n,k}$. Using~the relation between dual potentials and the EMOT plan from \eqref{eq:opt_plan}, we further obtain a neural proxy of~$\pi_\epsilon$:
\begin{equation}\label{eq:neural_plan}    
    d\pi_\epsilon^{\bm{\theta}^\star}\coloneqq \exp\left(\frac{\bigoplus_{i=1}^k f_{\theta^\star_i}-c}{\epsilon}\right)d(\otimes_{i=1}^k\mu_i).
\end{equation}
As elaborated in Section \ref{sec:complexity}, the NEMOT framework transfers the computational complexity from the dataset size $n$ to the mini-batch size $b$, which is typically orders of magnitude smaller than $n$.


\subsection{Performance Guarantees}
We provide non-asymptotic bounds on the NEMOT estimation error, both for the cost and the induced neural plan. Generally, the proposed approach entails three sources of error: (i) function approximation of the dual MOT potentials by neural nets; (ii) empirical estimation of the means by sample averages; and (iii) optimization, which comes from employing suboptimal (e.g., gradient-based) routines. We provide sharp bounds on the errors of types (i) and (ii), leaving the account of the optimization error for future work. To simplify the presentation of our bound, we assume that $\cX_i\subseteq [-1,1]^d$, for all $i=1,\ldots,k$ (although our results readily extend to arbitrary compact spaces) and that the pairwise cost $\tilde{c}$ is continuously differentiable. Many transportation costs of interest satisfy this assumption, e.g., the $p$-Wasserstein cost for $p>1$. The cost function $c$, regularization parameter $\epsilon>0$, and the population distributions $\muk$ are fixed throughout this section.

We analyze the NEMOT realized by an $m$-neuron shallow network. For any $a>0$, let~$\cF_{\mathsf{nn}}^{\mspace{1mu}m}(a)$ be the class of neural nets $f:\RR^d \rightarrow \mathbb{R}$ given by $f(z)=\sum\nolimits_{i=1}^m \beta_i \phi\left(\langle w_i, z\rangle+b_i\right)+\langle w_0, z\rangle + b_0$, with 
\begin{align*}
    \max_{1 \leq i \leq m}\|w_{i}\|_1 \vee |b_i| \leq 1,\ \ \max_{1 \leq i \leq m}|\beta_i| \leq \frac{a}{2m},\ \ |b_0|,\|w_0\|_1 \leq a
\end{align*}
parameter bounds, where $\phi(z)=z\vee0$ is the ReLU activation. We write  $\widehat{\MOT}_{c,\epsilon}^{m,a}$ for NEMOT \eqref{eq:ne_emot} with neural potentials in $\cF_{\mathsf{nn}}^{\mspace{1mu}m}(a)$, and present the following cost error bound. 
\begin{theorem}[NEMOT error bound]\label{thm:ne_erro}
    There exists a constant $C>0$ depending only on $d\coloneqq\max_{i=1,\dots,k}d_i$ such that setting $a=C(1+\epsilon^{-\lfloor\frac{d}{2}\rfloor+3)})$, we have
    \begin{align}
    &\EE\left[\left| \widehat{\MOT}_{c,\epsilon}^{m,a}(\mathbf{X}^{n,k}) -  \MOT_{c,\epsilon}(\muk)\right|\right]\lesssim_d(1+\epsilon^{-(\lfloor\frac{d}{2}\rfloor+1)})k^{p(\cG_c)}(m^{-\frac{1}{2}}+n^{-\frac{1}{2}}),\label{eq:nemot_est_bound}
\end{align}
where $p(\cG_c)=d+6$ when $\mathsf{deg}(\cG_c)=O(k)$, and $p(\cG_c)=1$  otherwise.
\end{theorem}

Evidently, equating $n\asymp m$, the above bound yields the $n^{-1/2}$ parametric convergence rate, which is sharp. This implies minimax optimality of NEMOT for estimation of the EMOT cost over the class of compactly supported distributions as above. Theorem \ref{thm:ne_erro} is proven in Appendix \ref{proof:ne_error} by decomposing the error analysis into its approximation and estimation parts. The former is controlled by showing that the optimal dual potentials belong to a H\"older class of arbitrarily large smoothness, whence we may appeal to approximation error bounds from \cite{klusowski2018approximation}. For the estimation error, we use standard maximal inequalities from empirical process theory (see \cite{wang2024neural} for the bimarginal case). 
Theorem \ref{thm:ne_erro} readily extends to sigmoidal networks, with minor adjustments, by leveraging the results of \cite{barron1993universal}.




By \eqref{eq:neural_plan}, an optimized NEMOT induces a neural estimate $\pi_\epsilon^{\bm{\theta}^\star}$ of the EMOT plan $\pi_\epsilon$. To complement the cost error bound above, we next provide guarantees on the quality of this plan approximation in KL divergence. The result shows that the plan error is dominated by the cost error, up to a $1/\epsilon$ factor.

\begin{theorem}[NEMOT plan approximation]\label{thm:ne_plan_approx}
    Let $\bm{\theta}^\star$ be a maximizer of the NEMOT objective \eqref{eq:ne_emot} and consider the induced plan $\pi_\epsilon^{\bm{\theta}^\star}$ from \eqref{eq:neural_plan}. Denote the EMOT plan for $ \MOT_{c,\epsilon}(\muk)$ by $\pi_\epsilon^\star$, and write $\delta_{\mathsf{NEMOT}}$ for the NEMOT cost error bound on the right-hand side of \eqref{eq:nemot_est_bound}. We have \[  \DKL\big(\pi_\epsilon^{\star}\big\|\pi_\epsilon^{\bm{\theta}^\star}\big)\lesssim_d\epsilon^{-1}\delta_{\mathsf{NEMOT}}.
    \]
\end{theorem}

Inserting the expression for $\delta_{\mathsf{NEMOT}}$ with $n\asymp m$ yields $O(n^{-1/2})$ convergence for the plan estimation KL divergence error. The theorem is proven in Appendix \ref{proof:ne_plan_approx}.

\subsection{Computational Complexity of NEMOT}\label{sec:complexity}

\begin{wraptable}{r}{0.42\textwidth}   
  \vspace{-12pt}                        
  \centering
  \small                                
  \begin{tabular}{c|c|c}
    \hline
    \textbf{Cost-Graph} & \textbf{Sinkhorn} & \textbf{NEMOT} \\ \hline
    Full   & $O(n^k)$      & $O(nk^2)$ \\ 
    Circle & $O(n^3k^2)$   & $O(nk)$   \\ 
    Tree   & $O(n^2k)$     & $O(nk)$   \\ \hline
  \end{tabular}
  \caption{Computational complexity of cost evaluation for different cost graphs.}
  \label{tab:complexity_table}
\end{wraptable}

NEMOT is trained via mini-batch gradient-based optimization, which enables running it in regimes where the multimarginal Sinkhorn algorithm is infeasible. The bottleneck for Sinkhorn is evident already at the level of a single iteration, when the dataset $n$ and the number of marginals $k$ are large. Specifically, a single update of the iterated dual variables requires a marginalization of the $k$-dimensional coupling tensor---an operation whose complexity is generally $O(n^k)$ \cite{lin2022complexity}. A single epoch of NEMOT, on the other hand, iteratively computes the dual loss in \eqref{eq:ne_emot} and a parameter update via backpropagation. Since the objective can be computed on mini-batches, given a batch size $b$, the complexity of a single evaluation is $O(bk^2)$, which amounts to $O(nk^2)$ for the epoch. The $k^2$ term comes from the number of pairwise cost terms in $\cG_c$, and becomes linear when $\cG_c$ is sparse (e.g., circle or tree). This is a substantial decrease from the $O(n^k)$ per-iteration complexity of the multimarginal Sinkhorn. Even under a simple case of a tree cost graph, Sinkhorn’s complexity is $O(n^2k)$, which is still more prohibitive than the  $O(nk^2)$ per epoch of NEMOT with a full cost.

As mentioned, the optimization of NEMOT also employs backpropagation, whose computational cost we discuss next. For $k$ neural nets parameterizing the EMOT dual potentials, each with $L$ layers and at most $N$ parameters in each layer, the complexity of backpropagation is $O(nkLN^2)$ per epoch. In \cref{sec:numerical results}, we show that this term is potentially bigger than the NEMOT complexity, resulting in a decrease in average epoch time when $b$ increases. Again, the linear dependence on $n$ is a notable improvement compared to the $O(n^k)$ complexity of Sinkhorn iterations. The full run time of the two algorithms arises from iterating over the steps discussed above. We provide a numerical comparison in Section 5, where the speedup and increased scalability of NEMOT are clearly evident. In practice, as $n$ and $k$ grow (e.g., $n=1000$ and $k=4$, as seen in \cref{sec:numerical results}), Sinkhorn falters and cannot complete a single iteration, while NEMOT continues to provide accurate estimates.

\begin{remark}[NEMOT stability for small $n$]
    In the regime of small $n$, NEMOT can present instability and lead to erroneous results.
    Such instability stems from the sum of exponential terms in \eqref{eq:ne_emot}.
    In this case, one may replace this term with a $k$-sample $U$-statistic \cite{lee2019u}, which considers all $b^k$ possible $k$-tuples from a given sample batch.
    While incurring exponential complexity of $b^k$, incorporating a $U$-statistic results in the minimum-variance estimator of the exponential term.
    For sparse cost graph, the $U$-statistic complexity can be reduced to  polynomial complexity in $(b,k)$ using tools similar to the ones developed for Sinkhorn.
    We refer the reader to Appendix \ref{appendix:u-stat} for more details.
\end{remark}




\section{Neural Multimarginal Gromov-Wasserstein}
The Gromov-Wasserstein (GW) problem \cite{memoli2011gromov} provides a OT-based framework for alignment of heterogeneous datasets (e.g., of varying modalities or semantics). Representing each dataset as a metric measure (mm) spaces, say $(\cX,\mathsf{d},\mu)$, we look for an alignment plan that optimally matches the mm spaces' internal structures by minimizing distance distortion. GW alignment has seen a variety of applications, including single-cell genomics~\cite{cao2022manifold, demetci2022scot}, matching of language models~\cite{alvarez2018gromov}, and shape/graph matching~\cite{memoli2011spectral}, with recent interest in the multimarginal extension of the problem~\cite{beier2023multi}. We propose utilizing NEMOT as a primitive for a neural entropic MGW (NEMGW) algorithm. 


We consider the quadratic entropic MGW (EMGW) problem between $k$ Euclidean mm spaces $(\cX_1,\mathsf{d}_1,\mu_1),\ldots,(\cX_k,\mathsf{d}_k,\mu_k)$ of (possibly) different dimensions, i.e., $\cX_i\subseteq \RR^{d_i}$ for $i=1,\ldots,k$, and $\mathsf{d}_i(x,y)=\|x-y\|_2$ in $\RR^{d_i}$. The EMGW problem with parameter $\epsilon>0$ is 
\begin{align}
&\EMGW(\muk)\coloneqq \inf_{\pi \in \Pi(\muk)}\int \Delta(\bxk,\byk)  d\pi\otimes\pi(\bxk,\byk)+ \epsilon\DKL(\pi\|\muk),\label{eq:EMGW} 
\end{align}
where $\Delta(\bxk,\byk)\coloneqq \sum_{(i,j)\in\cE} \left|\mathsf{d}_{i}(x_{i},y_{i})^2-\mathsf{d}_{j}(x_{j},y_{j})^2\right|^2$ is the multimarginal distance distortion cost, with $\cE$ an edge set specifying the connectivity profile. Finiteness of this distance requires marginals with finite 4th absolute moments, i.e., $\int_{\cX_i}\|x\|^4d\mu_i(x)<\infty\ \ \forall i=1,\ldots,n$, which we assume throughout. The unregularized multimarginal GW problem is $\MGW(\muk)\coloneqq \MGW_0(\muk)$.

\paragraph{Entropic approximation gap.} Entropic GW is a popular computationally tractable proxy of the unregularized GW, which is an instance of the NP-hard quadratic assignment problem \cite{Commander2005}. However, a formal justification for this approximation was provided only recently as an $O(\epsilon\log(1/\epsilon))$ bound on the entropic GW cost gap \cite{zhang2024gromov}. We extend this result to MGW via an argument that combines a block approximation technique of optimal plans and the Gaussian maximum entropy inequality.


\begin{proposition}[Entropic gap]\label{prop:mgw_cost_approx}
    For any $\epsilon\in(0,1]$, $\EMGW(\muk) - \MGW(\muk) \lesssim_d k\epsilon \log \left(\frac{|\cE|}{k\epsilon}\right)$.
\end{proposition}

\paragraph{Variational form.} To leverage NEMOT for EMGW computation, we derive a new variational representation of EMGW that rewrites it as an infimum of a certain class of EMOT problems. This generalizes the corresponding representation in the bimarginal case that was recently derived in \cite{zhang2024gromov}. Assume, without loss of generality, that the population measures $\mu_1,\ldots,\mu_k$ are all centered (EMGW is invariant to translations). By expanding the quadratic in the EMGW cost $\Delta$ from \eqref{eq:EMGW} and separating terms that depend on $\pi$, we may decomposition $\EMGW(\muk) = \sS^1(\muk) + \sS^2_\epsilon(\muk)$, where 
\begin{align}
    \sS^1(\muk)&\coloneqq \sum_{(i,j)\in\cE}\sum_{\ell=i,j} \int \|x_{\ell}-y_{\ell}\|^4 d\mu_{\ell}\otimes \mu_{\ell}(x_{\ell},y_\ell)-4\int\| x_{i}\|^2\|x_{j} \|^2d\mu_{i}\otimes\mu_j(x_i,x_{j})\label{eq:mgw_s1}\\
    \sS^2_\epsilon(\muk)&\coloneqq \inf_\pi \sum_{(i,j)\in\cE} -4\int \|x_{i}\|^2\|x_{j}\|^2d\pi_{ij}(x_{i},x_j) -8\int\langle x_{i},y_{i} \rangle \langle x_j,y_j \rangle d\pi_{ij}^{\otimes 2}(x_{i},y_i,x_{j},y_{j})\nonumber\\
&\mspace{480mu}+\epsilon\DKL(\pi\|\muk),\label{eq:mgw_s2}
\end{align}
with $\pi_{ij}$ as the $(i,j)$th marginal of $\pi\in\cP(\cxk)$. 

We derive a variational form for $\sS^2_\epsilon(\muk)$ that represents it as an infimum of a class of EMOT problems. Write $M_2(\mu)\coloneqq\int \|x\|^2d\mu(x)$ and set $M_{ij}=\sqrt{M_2(\mu_i)M_2(\mu_j)}$ for $i,j=1,\ldots,k$. Further, given the cost edge set $\cE$ from \eqref{eq:EMGW} and $|\cE|$ matrices $\rA_{ij}\in\RR^{d_i\times d_j}$, $(i,j)\in\cE$, write $\mathbf{A}_\cE=\{\rA_{ij}\}_{(i,j)\in\cE}$ and define $\|\mathbf{A}_\cE\|_\rF\coloneqq\big(\sum_{(i,j)\in\cE} 32 \|\rA_{ij}\|^2_\rF\big)^{1/2}$, where $\|\cdot\|_\rF $ is the Frobenius~norm. The following result is derived by extending the bimarginal argument from \cite[Theorem 1]{zhang2024gromov}  (see Appendix \ref{appendix:emgw_lin}).

\begin{proposition}[Variational form]\label{prop:mgw_s2}
    The following holds
    \begin{equation}
    \sS^2_\epsilon(\muk) = \inf_{\mathbf{A}_\cE}\ \ 32 \|\mathbf{A}_\cE\|^2_\rF  + \MOT_{\mathbf{A}_{\cE},\epsilon}(\muk),\label{eq:s2_lin}
    \end{equation}
where $\MOT_{\mathbf{A}_\cE,\epsilon}$ is EMOT with cost function $c_{\mathbf{A}_\cE}(\bxk)\coloneqq \sum_{(i,j)\in\cE} \left(-4 \|x_{i}\|^2\|x_{j}\|^2 + 32x_{i}^\intercal \rA_{ij} x_{j}\right)$. In addition, the infimum is achieved inside the compact set where $\|\rA_{ij}\|_\rF\leq M_{ij}/2$, for all $(i,j)\in\cE$.  


\end{proposition}
This variational form links the EMGW problem and EMOT, for which we have the neural estimator. This suggests computing EMGW via an alternating optimization over the $\mathbf{A}_\cE$ auxiliary matrices and the NEMOT parameters for $\MOT_{\mathbf{A}_\cE,\epsilon}(\muk)$. We explore this approach next. 


\textbf{Neural EMGW via Alternating Optimization} Given a dataset $\mathbf{X}^{n,k}$ with $n$ samples from each of $k$ mm spaces, we propose the following neural EMGW (NEMGW) estimator for $\EMGW(\muk)$:
\begin{align}
    &\widehat{\MGW}_\epsilon(\textbf{X}^{n\times k}) \coloneqq\hat{\sS}^1_\epsilon(\textbf{X}^{n\times k})+\inf_{\mathbf{A}_{\cE}}\left(\|\mathbf{A}_{\cE}\|^2_\mathrm{F}+\widehat{\MOT}_{\mathbf{A}_{\cE},\epsilon}(\textbf{X}^{n\times k})\right),\label{eq:nemgw}
\end{align}
where $\hat{\sS}^1_\epsilon(\textbf{X}^{n\times k})$ is a sample-mean estimate
of \eqref{eq:mgw_s1}. Calculating \eqref{eq:nemgw} entails alternating between NEMOT and $\mathbf{A}_{\cE}$ optimization. For NEMOT, a fixed $\mathbf{A}_{\cE}$ induces a cost $c_{\mathbf{A}_\cE}$, for which we can apply Algorithm \ref{alg:nemot}. To optimize over $\mathbf{A}_\cE$ we may adopt a first-order method, as described next.

Denote $\Phi_\epsilon({\mathbf{A}_\cE},\pi)=
32 \|\mathbf{A}_\cE\|^2_\rF  +\int_{\cxk}c_{\mathbf{A}_\cE}(\bxk)d\pi(\bxk)$ and observe that infimizing this objective over $\mathbf{A}_\cE$ and $\pi\in\Pi(\muk)$ retrieves the $\sS^2_\epsilon(\muk)$ part in the decomposition of EMGW (see \cref{prop:mgw_s2}). For a fixed $\pi$, $\Phi_\epsilon(\cdot,\pi)$ is convex in $\mathbf{A}_{\cE}$, and we may find the global minimum using gradient methods. This observation was made in \cite{rioux2024entropic}, leading to the first efficient and provably convergent (with rates) algorithms for computing bimarginal EGW. The scheme alternates between gradient steps and Sinkhorn iterations, and is the inspiration for our approach. To that end, we generalize the gradient analysis to the multimarginal setting and replace the Sinkhorn step with NEMOT.



In practice, we only have access to the samples $\mathbf{X}^{n,k}$, and the populations  $\mu_1,\ldots,\mu_k$ are replaced with their empirical measures $\hat\mu_\ell^n\coloneqq\frac{1}{n}\sum_{j=1}^n \delta_{X_{i\ell}}$, $\ell=1,\ldots,k$. In this (discrete) case, the plan $\pi$ is identified with a tensor $\mathbf{\Pi}\in\RR^{d_1\times\ldots\times d_k}$. The gradient of $\Phi_\epsilon(\cdot,\mathbf{\Pi})$ w.r.t. each $\rA_{ij}$ is now given by:
\begin{equation}\label{eq:A_update_emgw}
    \nabla_{\rA_{ij}}\Phi_\epsilon(\mathbf{A}_\cE,\mathbf{\Pi}) = 64\rA_{ij} -32 \textbf{X}_i\Pi_{ij} \textbf{X}_j^\intercal,
\end{equation}
where $\Pi_{ij}$ is the marginalization of $\mathbf{\Pi}$ to the $(i,j)$th pair of spaces, i.e., $\Pi_{ij}[\ell_i,\ell_j]=\sum_{\{\ell_1,\dots,\ell_k\}/\{\ell_i,\ell_j\}}\mathbf{\Pi}[\ell_1,\dots,\ell_k]$, where $\mathbf{\Pi}[\,\cdot\,]$ denotes an entry of the tensor. Evolving $\mathbf{A}_{\cE}$ along the gradient will converge towards the global solution $\mathbf{A}_\cE^\star$, for the prescribed $\mathbf{\Pi}$. We note that the convergence guarantees in Theorems~9 and~11 in \cite{rioux2024entropic} readily extend to the multimarginal case and apply in our setting.

To obtain the plan tensor $\mathbf{\Pi}$ we use NEMOT, as optimized neural potentials $(f_{\theta_i})_{i=1}^k$ induce a good approximation of the ground truth plan via \eqref{eq:neural_plan} and \cref{thm:ne_plan_approx}. The gradient update for $\mathbf{A}_{\cE}$ is alternated with the NEMOT \cref{alg:nemot} to obtain the final NEMGW estimate $\widehat{\MGW}_\epsilon(\textbf{X}^{n\times k})$. The approach is independent of the specific solver employed to obtain the plan. For small $n$ and/or $k$ values, it typically suffices to use the Sinkhorn algorithm, as in \cite{rioux2024entropic}. In the neural estimation setting, however, it is convenient to utilize automatic differentiation packages to compute the gradient in $\mathbf{A}_\cE$. We can then incorporate its optimization into the overall routine that also optimizes the NEMOT parameters $\bm{\theta}$.
Combining the NEMOT and NEMGW further enables neural estimation of the \emph{fused EMGW distance} \cite{vayer2020fused}, which gained recent attention for matching of structured, labeled data.


\begin{figure*}[!t]
    \centering
    \subfigure[Uniform distributions]
    {
        \scalebox{0.7}{
\begin{tikzpicture}

\definecolor{chocolate2267451}{RGB}{226,74,51}
\definecolor{dimgray85}{RGB}{85,85,85}
\definecolor{gainsboro229}{RGB}{229,229,229}
\definecolor{steelblue52138189}{RGB}{52,138,189}
\definecolor{springgreen60179113}{RGB}{50,149,94}

\begin{axis}[
width=6.5cm, height=5cm,
legend style={at={(0.985,0.05)}, anchor=south east , fill opacity=0.65, draw opacity=1, text opacity=1, draw=none},
axis background/.style={fill=gainsboro229},
axis line style={white},
tick align=outside,
tick pos=left,
x grid style={white},
xlabel=\textcolor{dimgray85}{$d$},
xmajorgrids,
xmin=0.8, xmax=1300,
xmode=log,
xtick style={color=dimgray85},
y grid style={white},
ylabel=\textcolor{dimgray85}{OT loss},
ymajorgrids,
ymin=0.565, ymax=0.68,
ytick style={color=dimgray85}
]
\addplot [only marks, steelblue52138189 , mark=*, mark size=3, mark options={solid}]
table {%
1		0.58
2       0.622
5		0.64
10		0.646
15		0.652
25		0.653
50		0.655
100		0.658
150		0.6634
250		0.6634
500		0.6613
1000	0.6641
};
\addplot [semithick, chocolate2267451, mark=none, mark size=3, mark options={solid}, dashed]
table {%
1		0.58
2       0.62398
5		0.64
10		0.646
15		0.652
25		0.655
50		0.654
100		0.658
150		0.659
250		0.66
500		0.663
1000	0.665
};
\addlegendentry{NEMOT (Ours)}
\addlegendentry{Sinkhorn}

\end{axis}

\end{tikzpicture}}
        \label{fig:nemot_est_unif}
    }
    \hfill
    \subfigure[Gaussian distributions]
    {
        \scalebox{0.7}{
\begin{tikzpicture}

\definecolor{chocolate2267451}{RGB}{226,74,51}
\definecolor{dimgray85}{RGB}{85,85,85}
\definecolor{gainsboro229}{RGB}{229,229,229}
\definecolor{steelblue52138189}{RGB}{52,138,189}
\definecolor{springgreen60179113}{RGB}{50,149,94}

\begin{axis}[
width=6.5cm, height=5cm,
legend style={at={(0.985,0.05)}, anchor=south east , fill opacity=0.65, draw opacity=1, text opacity=1, draw=none},
axis background/.style={fill=gainsboro229},
axis line style={white},
tick align=outside,
tick pos=left,
x grid style={white},
xmode=log,
xlabel=\textcolor{dimgray85}{$d$},
xmajorgrids,
xmin=-0.2, xmax=1450,
xtick style={color=dimgray85},
y grid style={white},
ymajorgrids,
ymin=1.2, ymax=2.05,
ytick style={color=dimgray85}
]
\addplot [only marks, steelblue52138189, mark=*, mark size=3, mark options={solid}]
table {%
1 1.29177
2 1.63466
5 1.86109
8 1.89998
10 1.92921
15 1.95815
20 1.96529
25 1.97345
50 1.9939
100 1.99804
200 1.99956
300 1.99854
500 2.00036
1000 2.00104
};
\addplot [semithick, chocolate2267451, mark=none, mark size=3, mark options={solid}, dashed]
table {%
1 1.28977
2 1.64034
5 1.85228
8 1.89673
10 1.92199
15 1.96226
20 1.96491
25 1.97667
50 1.98801
100 1.97942
200 1.99514
300 1.99875
500 2.00159
1000 2.00598
};
\addlegendentry{NEMOT (ours)}
\addlegendentry{Sinkhorn}
\end{axis}

\end{tikzpicture}}
        \label{fig:nemot_est_gauss}
    }
    \hfill
    \subfigure[GMM distributions]
    {
        \scalebox{0.7}{
\begin{tikzpicture}

\definecolor{chocolate2267451}{RGB}{226,74,51}
\definecolor{dimgray85}{RGB}{85,85,85}
\definecolor{gainsboro229}{RGB}{229,229,229}
\definecolor{steelblue52138189}{RGB}{52,138,189}
\definecolor{springgreen60179113}{RGB}{50,149,94}

\begin{axis}[
width=6.5cm, height=5cm,
legend style={at={(0.985,0.05)}, anchor=south east , fill opacity=0.65, draw opacity=1, text opacity=1, draw=none},
axis background/.style={fill=gainsboro229},
axis line style={white},
tick align=outside,
tick pos=left,
x grid style={white},
xlabel=\textcolor{dimgray85}{$d$},
xmajorgrids,
xmin=-0.2, xmax=55,
xtick style={color=dimgray85},
y grid style={white},
ymajorgrids,
ymin=0.8, ymax=3.3,
ytick style={color=dimgray85}
]
\addplot [only marks, steelblue52138189, mark=*, mark size=3, mark options={solid}]
table {%
3  1.0409640444565222
5  2.0431999548225646
8  2.073991778208192
10 2.127076572167139
15 2.2413241254296272
20 2.3374589411442956
25 2.4349545504763377
30 2.5456381961608385
35 2.6525895462610665
50 2.9543806840469884
};
\addplot [semithick, chocolate2267451, mark=none, mark size=3, mark options={solid}, dashed]
table {%
3  1.0533102258791902
5  2.040570434864882
8  2.092884640978608
10 2.137188383856879
15 2.2469485620558336
20 2.3378634732271086
25 2.449411800747849
30 2.5409641183244704
35 2.637924735434109
50 2.9378842652820527
};

\addlegendentry{NEMOT (ours)}
\addlegendentry{Sinkhorn}
\end{axis}

\end{tikzpicture}}
        \label{fig:gmm}
        
    }
    
    \caption{NEMOT estimation performance across different datasets. NEMOT consistently presents accurate estimates, benchmarked against multimarginal Sinkhorn. 
    }
    \label{fig:combined_figures_ne_err}
\end{figure*}
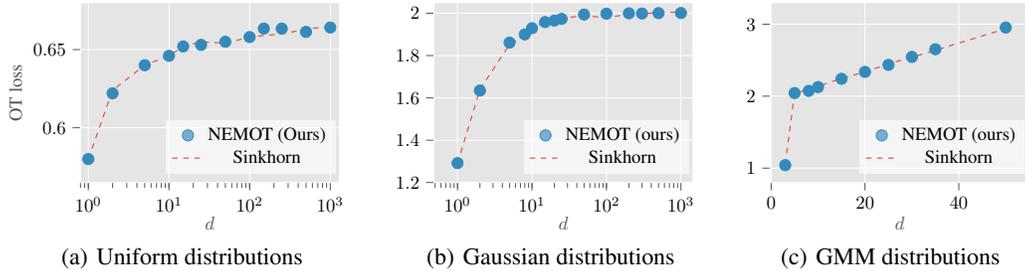

\begin{figure*}[!t]
    \centering
    \subfigure[Comparison: Full cost, $k=4$]
    {
        \scalebox{0.7}{
\begin{tikzpicture}

\definecolor{chocolate2267451}{RGB}{226,74,51}
\definecolor{dimgray85}{RGB}{85,85,85}
\definecolor{gainsboro229}{RGB}{229,229,229}
\definecolor{steelblue52138189}{RGB}{52,138,189}
\definecolor{springgreen60179113}{RGB}{50,149,94}

\begin{axis}[
width=7cm, height=5.5cm,
legend style={at={(1,0)}, anchor=south east , fill opacity=0.65, draw opacity=1, text opacity=1, draw=none},
axis background/.style={fill=gainsboro229},
axis line style={white},
log basis x={10},
log basis y={10},
tick align=outside,
tick pos=left,
x grid style={white},
xlabel=\textcolor{dimgray85}{$n$},
xmajorgrids,
xmin=38.3635249505463, xmax=13033.2132056306,
xmode=log,
xtick style={color=dimgray85},
xtick={1,10,100,1000,10000,100000,1000000},
xticklabels={
  \(\displaystyle {10^{0}}\),
  \(\displaystyle {10^{1}}\),
  \(\displaystyle {10^{2}}\),
  \(\displaystyle {10^{3}}\),
  \(\displaystyle {10^{4}}\),
  \(\displaystyle {10^{5}}\),
  \(\displaystyle {10^{6}}\)
},
y grid style={white},
ylabel=\textcolor{dimgray85}{Runtime [sec]},
ymajorgrids,
ymin=6.02051610513617, ymax=7686.07140294662,
ymode=log,
ytick style={color=dimgray85},
ytick={0.1,1,10,100,1000,10000,100000},
yticklabels={
  \(\displaystyle {10^{-1}}\),
  \(\displaystyle {10^{0}}\),
  \(\displaystyle {10^{1}}\),
  \(\displaystyle {10^{2}}\),
  \(\displaystyle {10^{3}}\),
  \(\displaystyle {10^{4}}\),
  \(\displaystyle {10^{5}}\)
}
]
\addplot [semithick, chocolate2267451 , mark=*, mark size=3, mark options={solid}]
table {%
50 11.4376
100 172.2596
150 878.6094
175 1543.7124
200 2591.1846
250 5552.894
};
\addplot [semithick, black, mark=none, mark size=3, mark options={solid}, dashed,line width=1.5]
table {%
50				3.333333333333330
100				6.66666666666668
150				10.0
300				20.0
500				33.33333333333330
1000				66.6666666666668
5000				333.3333333333330
10000				666.666666666668
};
\addplot [semithick, steelblue52138189, mark=*, mark size=3, mark options={solid}]
table {%
50 8.85  
100 7.45  
150 7.65  
300 14.40  
500 20.45  
1000 42.75  
5000 190.60  
10000 494.00
};
\addlegendentry{\small Sinkhorn}
\addlegendentry{\small $O(n)$}
\addlegendentry{\small NEMOT}
\end{axis}

\end{tikzpicture}}
        \label{fig:runtime_fullcost}
    }
    \hfill
    \subfigure[Comparison: Circle cost, $k=10$]
    {
        \scalebox{0.7}{
\begin{tikzpicture}

\definecolor{chocolate2267451}{RGB}{226,74,51}
\definecolor{dimgray85}{RGB}{85,85,85}
\definecolor{gainsboro229}{RGB}{229,229,229}
\definecolor{steelblue52138189}{RGB}{52,138,189}
\definecolor{springgreen60179113}{RGB}{50,149,94}

\begin{axis}[
width=7cm, height=5.5cm,
legend style={at={(1,0)}, anchor=south east , fill opacity=0.65, draw opacity=1, text opacity=1, draw=none},
axis background/.style={fill=gainsboro229},
axis line style={white},
log basis x={10},
log basis y={10},
tick align=outside,
tick pos=left,
x grid style={white},
xlabel=\textcolor{dimgray85}{$n$},
xmajorgrids,
xmin=187.968011006144, xmax=99751.0156097103,
xmode=log,
xtick style={color=dimgray85},
xtick={10,100,1000,10000,100000,1000000},
xticklabels={
  \(\displaystyle {10^{1}}\),
  \(\displaystyle {10^{2}}\),
  \(\displaystyle {10^{3}}\),
  \(\displaystyle {10^{4}}\),
  \(\displaystyle {10^{5}}\),
  \(\displaystyle {10^{6}}\)
},
y grid style={white},
ymajorgrids,
ymin=8.74691133158223, ymax=228947.48299353,
ymode=log,
ytick style={color=dimgray85},
ytick={0.1,1,10,100,1000,10000,100000,1000000,10000000},
yticklabels={
  \(\displaystyle {10^{-1}}\),
  \(\displaystyle {10^{0}}\),
  \(\displaystyle {10^{1}}\),
  \(\displaystyle {10^{2}}\),
  \(\displaystyle {10^{3}}\),
  \(\displaystyle {10^{4}}\),
  \(\displaystyle {10^{5}}\),
  \(\displaystyle {10^{6}}\),
  \(\displaystyle {10^{7}}\)
}
]
\addplot [semithick, chocolate2267451, mark=*, mark size=3, mark options={solid}]
table {%
10000 144186
5000 28044
2500 7091
2000 3335
1000 2014
500 657
250 486
};
\addplot [semithick, black, mark=none, mark size=3, mark options={solid},dashed, line width=1.5]
table {%
75000 4166.66666666667
65000 3611.11111111111
50000 2777.77777777778
35000 1944.44444444444
25000 1388.88888888889
20000 1111.11111111111
15000 833.333333333333
10000 555.555555555556
5000 277.777777777778
2500 138.888888888889
2000 111.111111111111
1000 55.5555555555556
500 27.7777777777778
250 13.8888888888889
};
\addplot [semithick, steelblue52138189, mark=*, mark size=3, mark options={solid}]
table {%
75000 2784.611539
65000 2432.990341
50000 1890.331901
35000 1597.015086
25000 1348.599287
20000 1238.496041
15000 949.5969475
10000 642.6607108
5000 310.7453529
2500 165.3145065
2000 132.2776929
1000 64.74222462
500 33.37565106
250 17.98061873
};
\addlegendentry{\small Sinkhorn}
\addlegendentry{\small $O(n)$}
\addlegendentry{\small NEMOT}
\end{axis}

\end{tikzpicture}}
        \label{fig:runtime_circle}
    }
    \hfill
    \subfigure[Runtime vs. dimension.]
    {
        \scalebox{0.7}{
\begin{tikzpicture}

\definecolor{chocolate2267451}{RGB}{226,74,51}
\definecolor{dimgray85}{RGB}{85,85,85}
\definecolor{gainsboro229}{RGB}{229,229,229}
\definecolor{steelblue52138189}{RGB}{52,138,189}

\begin{axis}[
width=5.7cm, height=5.5cm,
legend style={at={(1,0.45)}, anchor=east , fill opacity=0.65, draw opacity=1, text opacity=1, draw=none},
axis background/.style={fill=gainsboro229},
axis line style={white},
tick align=outside,
tick pos=left,
x grid style={white},
xlabel=\textcolor{dimgray85}{$d$},
xmajorgrids,
xmode = log,
xmin=-0.2, xmax=1200,
xtick style={color=dimgray85},
y grid style={white},
ymajorgrids,
ymin=0.2, ymax=2.7,
ytick style={color=dimgray85}
]
\addplot [semithick, steelblue52138189, mark=*, mark size=3, mark options={solid}]
table {%
1 0.68261
5 0.476198
10 0.4788
25 0.4791
50 0.477
100 0.4759
250 0.479668
350 0.48056
500 0.4837
750 0.48005
1000 0.48107
};
\addplot [semithick, chocolate2267451, mark=*, mark size=3, mark options={solid}]
table {%
1 2.44
5 2.4545
10 2.5554
25 2.5249
50 2.43569
100 2.4225
250 2.4091
350 2.43314
500 2.4237
750 2.4281
1000 2.4359
};
\addlegendentry{NEMOT}
\addlegendentry{Sinkhorn}

\end{axis}

\end{tikzpicture}}
        \label{fig:runtime_v_d}
    }
    \caption{Comparison of NEMOT and Sinkhorn runtime vs. dataset size ($n$) and dimension ($d$). NEMOT shows significant runtime gains, meeting the $O(n)$ behavior expected from the theory (fixed $d=20$ shown); neither method is meaningfully affected by the data dimensionality (right).}
    \label{fig:combined_runtime}
\end{figure*}
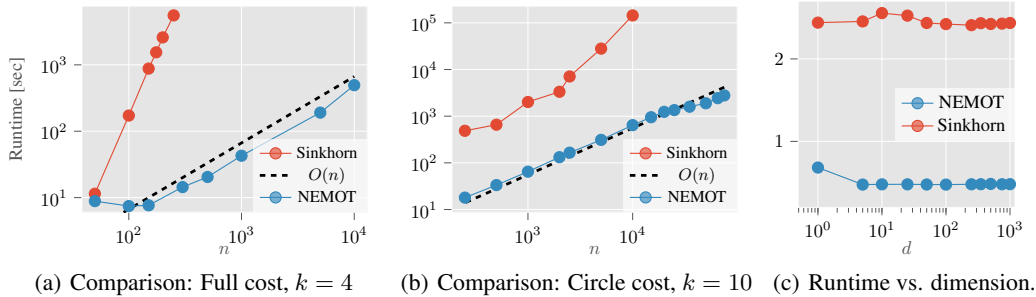

\section{Numerical Results.}\label{sec:numerical results}

In this section we empirically investigate our proposed NEMOT method in an array of synthetic data experiments. Our goal is to highlight the computational gains of NEMOT over the Sinkhorn algorithm, which is the prior state-of-the-art computational approach for EMOT. We opt not to pursue any specific applications of NEMOT as the application domain of MOT is broad and well-established \cite{cao2019multi,srivastava2018scalable,trillos2023multimarginal}. All results are averaged over $5$ seeds. Full implementation details are given in Appendix \ref{app:implementation}, and the code implementation can be found in the link \href{https://github.com/DorTsur/NMOT}{\texttt{https://github.com/DorTsur/NMOT}}.

\subsection{Comparison with Sinkhorn and Computational Gains}
\textbf{EMOT approximation.} 
We first verify that NEMOT successfully estimates the correct EMOT, for which we use the  multimarginal Sinkhorn algorithm \cite{lin2022complexity} as a baseline. We consider three settings for the $k$ population distributions: (i) uniform distributions over a $d$-dimensional normalized cube, i.e., 
$\mu_i=\mathsf{Unif}\left([-d^{-\frac{1}{2}},d^{-\frac{1}{2}}]^d\right)$; (ii) isotropic Gaussians $\mu_i=\cN(0,d^{-\frac{1}{2}}\rI_d)$ and (iii) Gaussian mixture models (see Appendix \ref{app:implementation} for sample construction procedure). In all cases, we compute the EMOT cost for a full-cost graph $\tilde{c}(\bxk)=\frac{1}{k}\sum_{i\neq j}\|x_i-x_j\|^2$. As shown in Figure \ref{fig:combined_figures_ne_err} NEMOT successfully estimates the underlying EMOT value for $d\in\{1,\dots,1000\}$.

\begin{figure*}[!t]
    \centering
    \subfigure[Average epoch time vs. batch size $b$, for several $k$ values, $n=10^4$, and $d=25$. Results for $k=3,4,5$ are shown in \textcolor{red}{red}, \textcolor{blue}{blue}, and black, respectively.]
    {
        \scalebox{0.72}{\begin{tikzpicture}

\definecolor{gainsboro229}{RGB}{229,229,229}

\begin{axis}[
width=6.5cm, height=5.5cm,
legend style={at={(0.95,0.95)}, anchor=north east, fill opacity=0.65, draw opacity=1, text opacity=1, draw=none},
axis background/.style={fill=gainsboro229},
axis line style={white},
xlabel={$b$},
ylabel={Average Epoch time [sec]},
xtick={0,1,2,3,4,5},
xticklabels={32,64,128,256,512,1028},
ymin=0, ymax=18,
bar width=6.5pt,
ymajorgrids
]
\addplot[ybar, fill=red,forget plot] coordinates {
    (-0.25, 7.33)
    (0.75, 3.78)
    (1.75, 1.9)
    (2.75, 1.05)
    (3.75, 0.62)
    (4.75, 0.27)
};

\addplot[ybar, fill=blue,forget plot] coordinates {
    (0.0, 12.1)
    (1.0, 5.59)
    (2.0, 3.07)
    (3.0, 1.76)
    (4.0, 0.81)
    (5.0, 0.44)
};

\addplot[ybar, fill=black, forget plot] coordinates {
    (0.25, 17.3)
    (1.25, 8.6)
    (2.25, 4.3)
    (3.25, 2.3)
    (4.25, 1.2)
    (5.25, 0.6)
};

\end{axis}
\end{tikzpicture}}
        \label{fig:time_vs_b}
    }
    \hfill
    \subfigure[Average epoch time vs. $k$ for two cost  structures and $n=5000$. \textcolor{red}{Red} line shows the practical limit on $k$ for full-cost~Sinkhorn.] 
    {
        \scalebox{0.72}{
\begin{tikzpicture}

\definecolor{chocolate2267451}{RGB}{226,74,51}
\definecolor{dimgray85}{RGB}{85,85,85}
\definecolor{gainsboro229}{RGB}{229,229,229}
\definecolor{steelblue52138189}{RGB}{52,138,189}
\definecolor{springgreen60179113}{RGB}{50,149,94}

\begin{axis}[
width=6cm, height=5.5cm,
legend style={at={(1,0)}, anchor=south east , fill opacity=0.65, draw opacity=1, text opacity=1, draw=none},
axis background/.style={fill=gainsboro229},
axis line style={white},
log basis x={10},
log basis y={10},
tick align=outside,
tick pos=left,
x grid style={white},
xlabel=\textcolor{dimgray85}{$k$},
xmajorgrids,
xmin=0, xmax=52,
xtick style={color=dimgray85},
y grid style={white},
ymajorgrids,
ymin=0, ymax=340,
ytick style={color=dimgray85},
]
\addplot [semithick, steelblue52138189, mark=*, mark size=3, mark options={solid}]
table {%
30  333.37
25	253.26972744060400
23	209.87988007132500
21	188.55171128490100
19	158.88990938961500
17	129.97020654082300
15	96.03834073479690
13	82.68552916709870
11	57.79686435695210
9	41.993164621719300
7	28.025685360814800
5	15.980013141036000
3	7.225820526480680
};
\addplot [semithick, blue, mark=square*, mark size=3, mark options={solid}]
table {%
50			317.83401215566100
40			202.02485056157600
30			132.7932177820380
25			105.41222824177600
20			73.63943606764080
15			47.598524396973
10			19.574982104344000
8			12.777193884977300
5			5.238018376273770
3			2.5962364650198400
};
\addplot [line width=1.3pt, red, mark=none, mark size=3, mark options={solid}, dashed]
table {%
6			380
6			0
};
\addlegendentry{\small Full cost}
\addlegendentry{\small Circle cost}
\end{axis}

\end{tikzpicture}}
        \label{fig:time_vs_k}
    }
    \hfill
    \subfigure[Estimated EMOT vs. $k$, full cost graph, $d=100$, Uniform marginals.]
    {
        \scalebox{0.72}{
\begin{tikzpicture}

\definecolor{chocolate2267451}{RGB}{226,74,51}
\definecolor{dimgray85}{RGB}{85,85,85}
\definecolor{gainsboro229}{RGB}{229,229,229}
\definecolor{steelblue52138189}{RGB}{52,138,189}
\definecolor{springgreen60179113}{RGB}{50,149,94}

\begin{axis}[
width=5cm, height=5.5cm,
legend style={at={(0.985,0.05)}, anchor=south east , fill opacity=0.65, draw opacity=1, text opacity=1, draw=none},
axis background/.style={fill=gainsboro229},
axis line style={white},
tick align=outside,
tick pos=left,
x grid style={white},
xmode=log,
ymode= log,
xlabel=\textcolor{dimgray85}{$d$},
xmajorgrids,
xmin=0, xmax=110,
xtick style={color=dimgray85},
y grid style={white},
ylabel=\textcolor{dimgray85}{OT loss},
ymajorgrids,
ymin=0, ymax=50,
ytick style={color=dimgray85}
]
\addplot [steelblue52138189, mark=*, mark size=3, mark options={solid}]
table {%
100	33.0017
75	24.66848
50	16.3332
35	11.332343
25	8.000
20	6.3329
10	2.999490
8	2.3316
5	1.3320
3	0.664
};

\end{axis}

\end{tikzpicture}}
        \label{fig:est_vs_k}
    }
    \vspace{-0.2cm}
    \caption{NEMOT scalability analysis, unlocking new regimes of feasible $(b,k)$~values. }
    \label{fig:combined_figures_ablation}
    \vspace{-0.2cm}
\end{figure*}
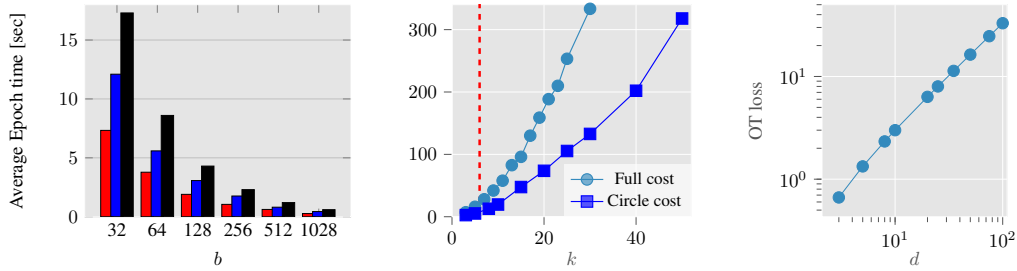

\textbf{Computational gains.}
Next, we demonstrate the computational speedup discussed in Section \ref{sec:complexity}.
We compare the overall runtime of both algorithms versus the dataset size. We first consider a full cost graph with $k=4$. As seen in Figure \ref{fig:runtime_fullcost}, NEMOT demonstrates linear time complexity, while the Sinkhorn algorithm shows intractability for $n>700$. We further note that solving full-cost Sinkhorn for $k>5$ is not feasible on standard machines (as also noted in \cite{piran2024contrasting}). 
For $k=10$ and a circle cost graph, therefore, we consider a circle cost graph, for which the Sinkhorn algorithm is more scalable. As seen in Figure \ref{fig:runtime_circle}, for $k=10$, NEMOT still introduces a speedup by orders of magnitude.

\textbf{Scalability.}
We explore the behavior of NEMOT w.r.t. two important parameters.
First, as applications of EMOT often consider it as a batch regularizer 
\cite{piran2024contrasting}, we investigate the NEMOT runtime when $b$ grows. As seen in Figure \ref{fig:time_vs_b}, for fixed $n=10^4$, the NEMOT epoch time does not pass 15 seconds. We note that the average epoch time decreases with the batch size. This is due to savings in the backpropagation, which is the majority of the epoch runtime.
Note that prior works only considered $b\leq 128$, as larger batch sizes resulted in large GPU usage, especially in higher dimensions.
Figure~\ref{fig:time_vs_k}, presents the average epoch time for a range of $k$ values, showing full-cost epoch times that are only a few minutes for even $k=30$, while Sinkhorn cannot be applied for $k >5$.
This implies that NEMOT enables comparison of an order of magnitude more marginals than was previously possible with the Sinkhorn algorithm (even after data downsampling). We provide an  estimate of the EMOT between a set of $k$ uniform variables for $3\leq k\leq 100$ (Figure \ref{fig:est_vs_k}). While a baseline is absent for large $k$, Figure \ref{fig:nemot_est_unif} shows that NEMOT correctly estimates the EMOT for $k=3$ and we expect a linear scaling of the EMOT due to the linear scaling of the normalized full cost in $k$.

\subsection{Real-World Data}

\begin{figure}[!ht]
    \centering
    \subfigure[NEMOT estimate vs. number of marginals - Cosine similarity.]{%
        \includegraphics[width=0.3\linewidth]{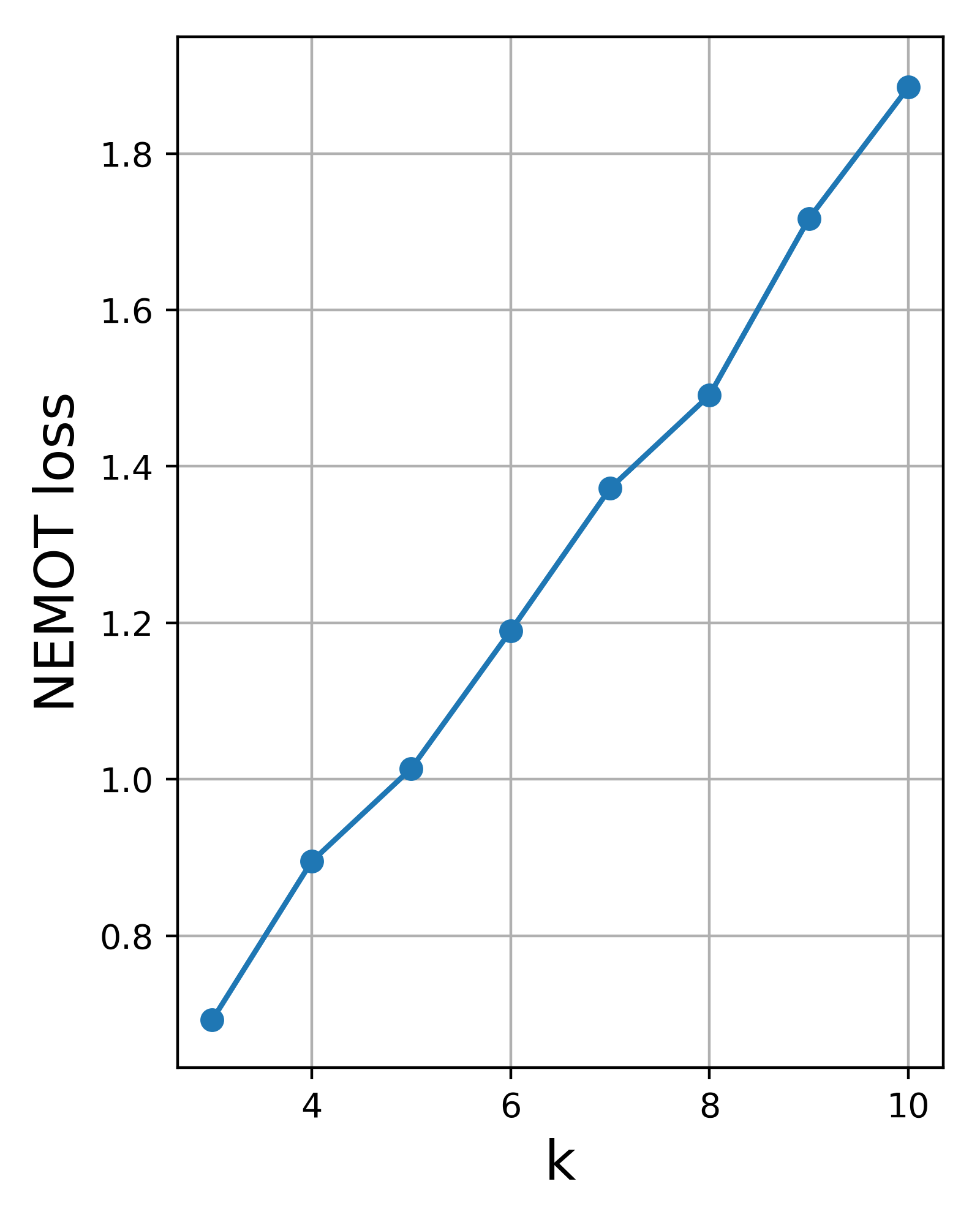}%
        \label{fig:cosim}%
    }\hfill
    \subfigure[NEMOT estimate vs. number of marginals - Squared Euclidean distance.]{%
        \includegraphics[width=0.3\linewidth]{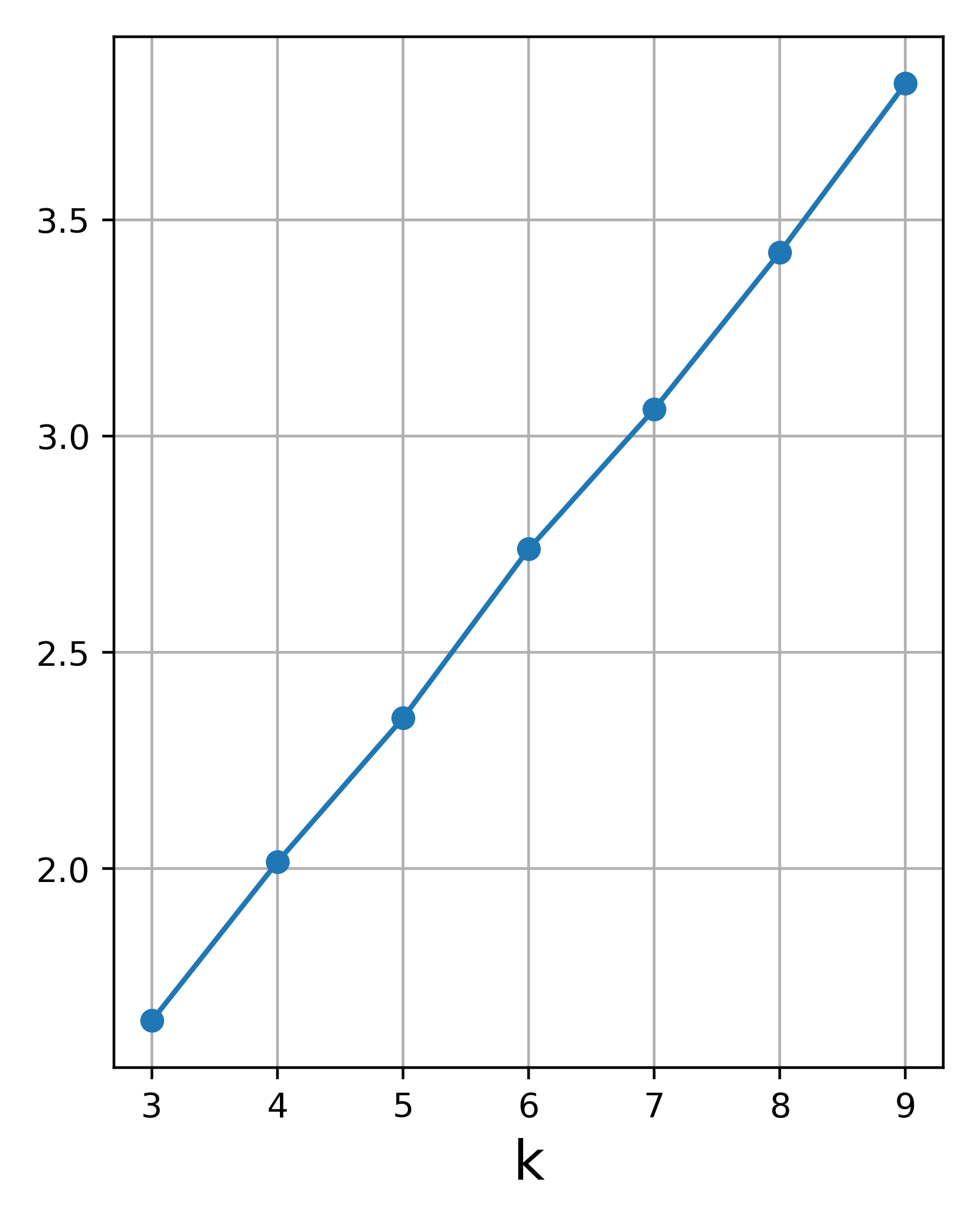}%
        \label{fig:mnist_vs_k}%
    }\hfill
    \subfigure[Average epoch time vs. number of marginals.]{%
        \includegraphics[width=0.3\linewidth]{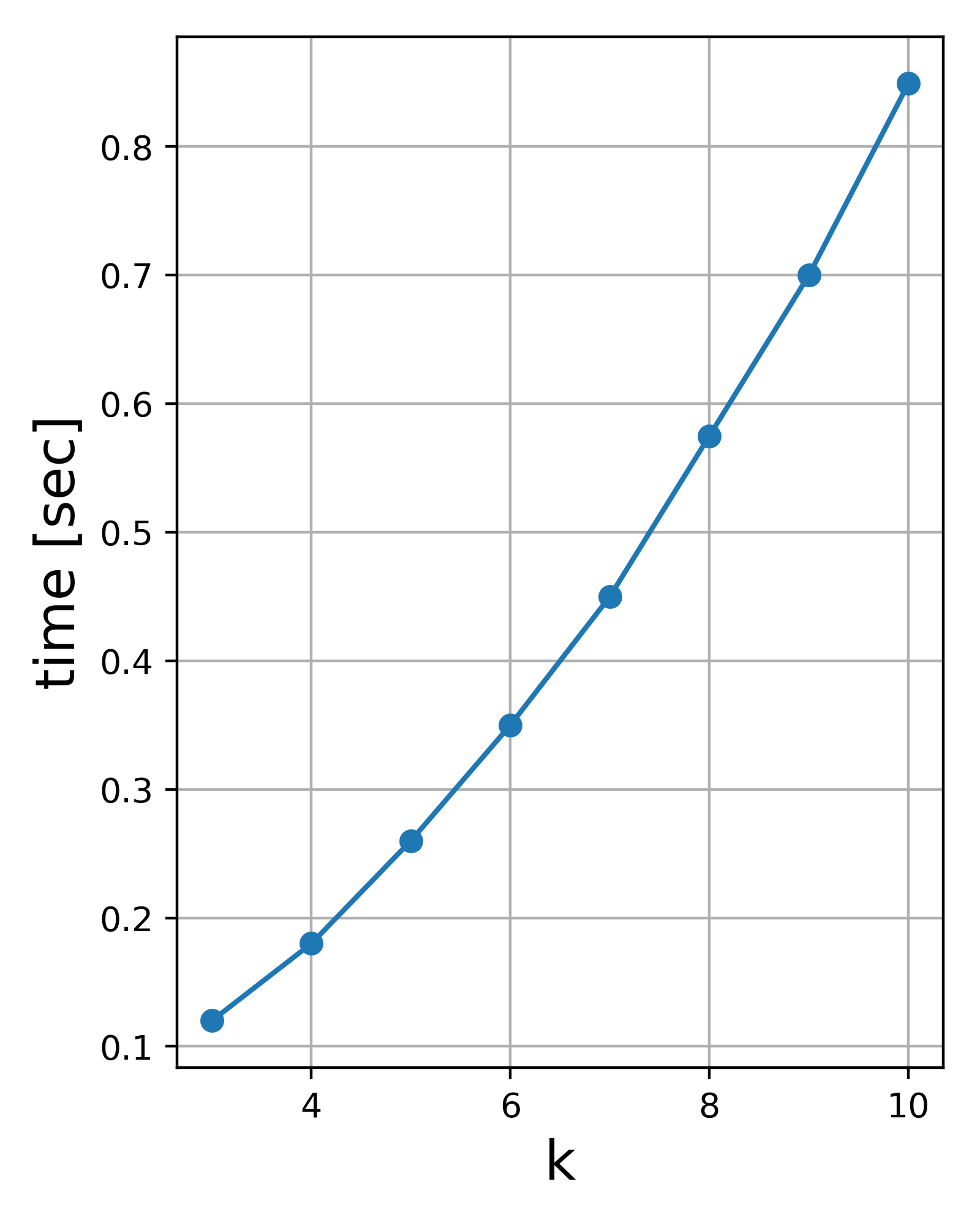}%
        \label{fig:cosim_t}%
    }
    \caption{NEMOT results - MNIST dataset.}
    \label{fig:mnist_cosim}
\end{figure}

\textbf{Estimation on MNIST data.} 
We demonstrate the performance of NEMOT on the MNIST dataset, which comprises $k=10$ classes of handwritten digit images. Each class is considered as a marginal.
We present the NEMOT estimate versus $k$, for the full cost graph comprising all pairwise Euclidean distances. Note that both the dataset size $(\sim 6\times10^3)$ and number of marginals are not feasible for the Sinkhorn algorithm. As seen in Figure~\ref{fig:mnist_vs_k}, the estimated EMOT grows linearly with the number of marginals. This is expected due to the square dependence on $k$ for the full cost graph and the normalization by $k^{-1}$.
Next, replace the Euclidean pairwise distance in the full cost graph with cosine similarity, which is a popular quantitative means of comparison between images and their features. 
We set the entropic regularization parameter to $\epsilon=0.1$ and apply the cosine similarity to the vectorized forms of the MNIST images. Figure \ref{fig:mnist_cosim}(a) shows the estimated NEMOT loss versus $k$, averaged over $5$ runs. We note that the estimated EMOT is linear with $k$, as expected. Figure \ref{fig:mnist_cosim}(b) illustrates the runtime versus $k$, where we note that NEMOT demonstrated similar behavior w.r.t. number of marginals, as observed for synthetic data. As evident from these experiments, the NEMOT framework provides efficient estimates on real-world data.


\begin{figure}[!b]
    \centering
    \includegraphics[width=0.8\linewidth]{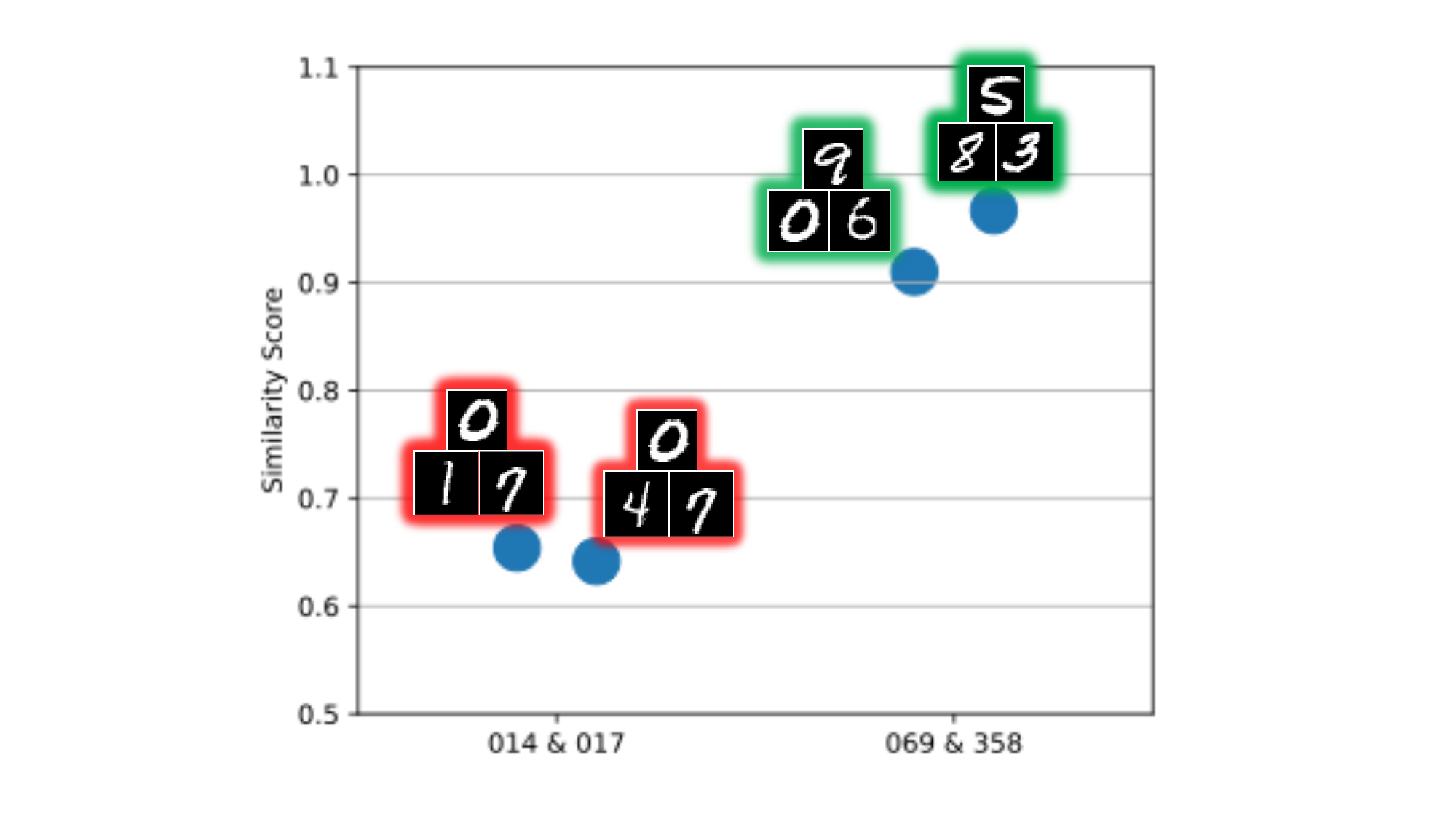}
    \caption{NEMOT-based similarity score between MNIST image datasets.}
    \label{fig:mnist_similary}
\end{figure}

\textbf{Application to evaluating perceptual similarity.} 
As the EMOT is a proxy for a matching loss \cite{piran2024contrasting}, we use it to evaluate the similarity between datasets based on some similarity score between their elements.
This corresponds to estimating an MOT-based matching score between the underlying distributions of the datasets .
As in our previous experiment, we divide the MNIST dataset into $k=10$ subsets, each corresponding to a digit (label). 
We expect subsets corresponding to perceptually similar digits to yield a higher EMOT value.
For the similar digits, we consider the sets $(3,5,8)$ and $(0,6,9)$, while the dissimilar ones are 
$(0,1,7)$ and $(0,1,4)$.
We use a full cost graph with pairwise cosine similarity. 
As seen in Figure \ref{fig:mnist_similary}, NEMOT consistently provides higher scores to perceptually similar datasets, while providing lower scores for dissimilar ones.

\begin{figure}[!t]
    \centering
    \subfigure[NEMOT estimate vs. number of marginals.]{%
        \includegraphics[width=0.48\linewidth]{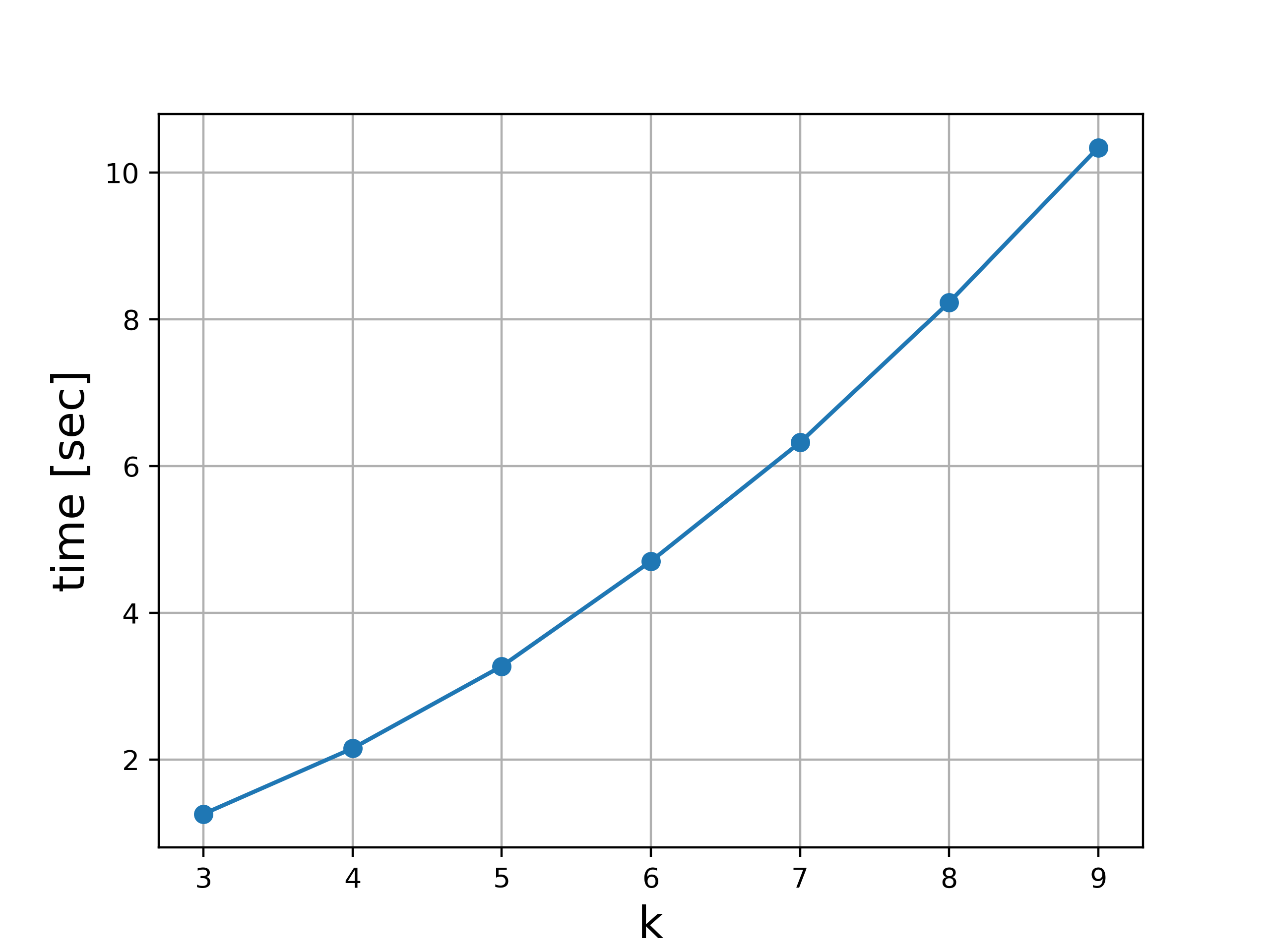}%
        \label{fig:cifar_nemot}%
    }\hfill
    \subfigure[Average epoch time vs. number of marginals.]{%
        \includegraphics[width=0.48\linewidth]{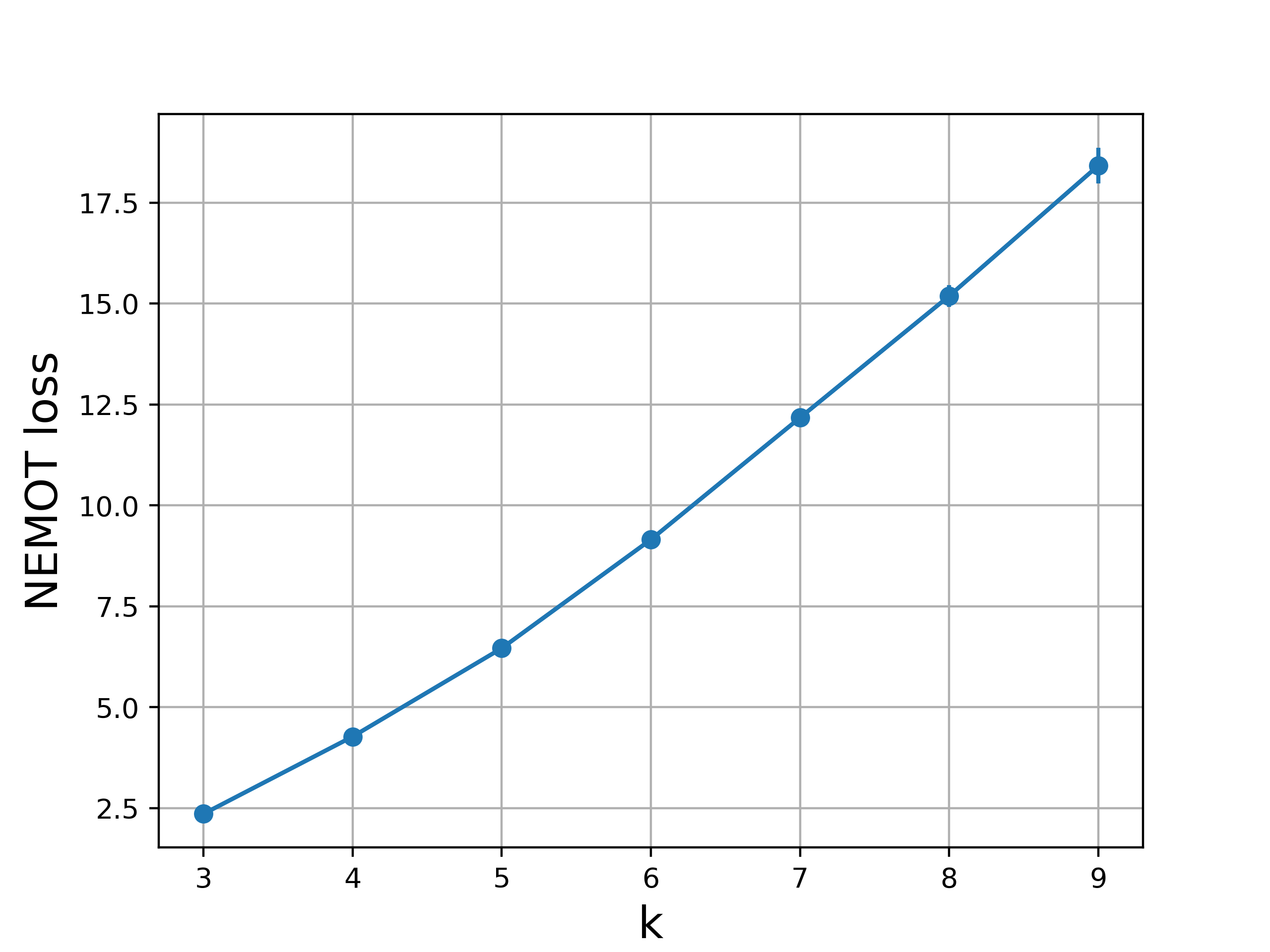}%
        \label{fig:cifar_t}%
    }
    \caption{CIFAR dataset, ResNet18-based cost.}
    \label{fig:cifar_retsuls}
\end{figure}

\textbf{Incorporating a deep-feature cost for CIFAR data.} 
We consider the CIFAR-10 dataset, which contains $5\times 10^4$ color images, divided into $10$ categories. Each image is of size $32\times 32$ and has $3$ channels.
To process this data, we implement the NEMOT with convolutional networks, whose architecture is delineated in 
\cref{tab:net_eot_cifar}.
We employ a neural network-based cost as the pairwise similarity metric for multi-channel images. This choice  also demonstrates the compatibility of NEMOT with less standard cost evaluations and complex pipelines.
Specifically, utilize a ResNet-18 model to extract deep features from the images, followed by a standard squared Euclidean distance between the extracted features. The ResNet-based pairwise costs are aggregated into a full-graph cost.

We report the estimated EMOT and average epoch time versus the number of marginals $k$, which correspond to the number of image classes considered.
Results are shown in Figure \ref{fig:cifar_retsuls}, where the linear dependence of the EMOT w.r.t. $k$ is maintained, as expected. 
Furthermore, Figure \ref{fig:cifar_t} shows that employing a CNN for the EMOT potentials and an additional ResNet for loss evaluation does not incur a significant overhead on the overall epoch time. 
We report the average epoch time as it is the temporal metric that is most relevant for applications that involve the NEMOT as a building-block.

\section{Conclusion}
This paper proposed NEMOT, an EMOT neural estimation algorithm utilizing modern neural network optimization schemes. NEMOT is applicable at scale with arbitrary cost functions and operates in a data-driven manner; these features align well with large-scale machine learning pipelines.
We derive formal guarantees on the NEMOT estimation error, for both the cost and optimal plan, under mild smoothness assumptions on the cost function. The resulting error bounds decay at the parametric rate, rendering NEMOT a minimax optimal estimator (provided an idealized optimizer, as the optimization error is not accounted for in our analysis).
Compared to the Sinkhorn algorithm---the prior go-to approach for EMOT computation---NEMOT offers a substantial speedup that becomes increasingly pronounced as the dataset size grows, while preserving comparable accuracy.
We further proposed to utilize NEMOT for a neural EMGW alignment solver. Due to the massive speedups achieved and increased practicality of the algorithm, we believe that NEMOT will unlock EMOT estimation and alignment for real-world datasets that were previously intractable.

\begin{table}[!ht]
\centering
\small
\begin{tabular}{@{}lll@{}}
\toprule
\textbf{Stage} & \textbf{Layer} & \textbf{Output shape} \\ \midrule
Input    & —                               & $(B,3,32,32)$   \\
Conv\,1  & $\mathrm{Conv}\,3{\times}3,\;3\!\rightarrow\!32$     & $(B,32,32,32)$ \\
Pool\,1  & $\mathrm{MaxPool}\,2{\times}2$          & $(B,32,16,16)$ \\
Conv\,2  & $\mathrm{Conv}\,3{\times}3,\;32\!\rightarrow\!64$    & $(B,64,16,16)$ \\
Pool\,2  & $\mathrm{MaxPool}\,2{\times}2$          & $(B,64,8,8)$   \\
Conv\,3  & $\mathrm{Conv}\,3{\times}3,\;64\!\rightarrow\!128$   & $(B,128,8,8)$  \\
Pool\,3  & $\mathrm{MaxPool}\,2{\times}2$          & $(B,128,4,4)$  \\
Flatten  & —                               & $(B,2048)$      \\
FC\,1    & $\mathrm{Linear}\;2048\!\rightarrow\!256$            & $(B,256)$      \\
Output   & $\mathrm{Linear}\;256\!\rightarrow\!1$               & $(B,1)$        \\ \bottomrule
\end{tabular}
\caption{convolutional NEMOT network architecture.}
\label{tab:net_eot_cifar}
\end{table}


\newpage

\bibliography{bibliography}
\bibliographystyle{unsrt}

\newpage
\appendix
\onecolumn

\begin{center}
    \textbf{\LARGE Supplementary material for:\\
    Neural Estimation for Scaling Entropic Multimarginal Optimal Transport}
\end{center}

\medskip
\medskip
\medskip
\medskip
\medskip
\medskip
\medskip
\medskip
\medskip

{
\large
\textbf{\underline{Table of contents:}}
}

\begin{itemize}
    \item \textbf{Section \ref{apdx:proofs}:} Proofs of the main theorems, propositions and lemmas.
    \item \textbf{Section \ref{app:implementation}:} Additional implementation details, including hypereparameter values, networks sizes, training routines and data generation schemes.
    \item \textbf{Section \ref{appendix:u-stat}:} A discussion on NEMOT implementation via U-statistics.
\end{itemize}
\newpage

\section{Proofs}\label{apdx:proofs}
\subsection{Proof of Theorem \ref{thm:ne_erro}}\label{proof:ne_error}
We begin by defining the population-level NEMOT as 
$$
\widehat{\MOT}^\epsilon_{m,a}(\muk)\sup_{f_{\theta_1},\dots,f_{\theta_k}\in\times_{\cF_{m_i,d_i}}}\sum_{i=1}^k\int f_{\theta_i} d\mu_i-\epsilon\int e^{\left(\bigoplus_{i=1}^k f_{\theta_i} -c\right)/\epsilon}d(\otimes_{i=1}^k\mu_i)+\epsilon.
$$
First, note divide the neural estimation error into two components
\begin{align*}
    &\EE\left[\left| \widehat{\MOT}^\epsilon_{m,a}(\mathbf{X}^{k\times n}) -  \MOT_{c,\epsilon}(\muk)\right|\right]\\
    &\hspace{3cm}\leq \underbrace{\EE\left[\left| \MOT^\epsilon_{c,m,a}(\muk) - \MOT_{c,\epsilon}(\muk) \right|\right]}_{\text{(approximation)}} + \underbrace{\EE\left[\left| \widehat{\MOT}^\epsilon_{m,a}(\mathsf{X}^{k\times n}) - \MOT^\epsilon_{c,m,a}(\muk) \right|\right]}_{\text{(estimation)}},
\end{align*}
where (approximation) quantifies the approximation error that is induced by replacing the optimal MOT potentials with elements from the class of shallow ReLU neural nets and (estimation) quantifies the error the one accumulates by replacing expectations with sample means. 

We next bound the approximation error. Denote the optimal set of potentials with $\left(\varphi_1^\star,\dots,\varphi_k^\star\right)$.\footnote{Existence of dual potentials is guaranteed under the assumption of finitely-supported measures \cite{carlier2020differential}.} The approximation error is upper bounded by
\begin{align*} 
    0&\leq \MOT_{c,\epsilon}(\muk)- \MOT^\epsilon_{c,m,a}(\muk)\\
    &\leq\sup_{f_{\theta_1},\dots,f_{\theta_k}\in\times_{\cF_{m_i,d_i}}}\sum_{i=1}^k\int_{\cX_i}\left(\varphi_i^\star-f_i\right)d\mu_i \\
    &\hspace{1cm}+ \int\left(\exp\left(\frac{\sum_{i=1}^k\varphi^\star_i(X_i)-c(X^k)}{\epsilon}\right)- \exp\left(\frac{\sum_{i=1}^k f_i(X_i)-c(X^k)}{\epsilon}\right)\right)d(\otimes_{i=1}^k \mu_i).
\end{align*}
Due to compactness of $\cX^k$, the exponential function is consequently Lipschitz continuous on $(-\infty,M_\cX]$ with Lipschitz constant $e^{M_{\cX^k}}$, where $M_\cX\coloneqq\sup_{x^k\in\cX^k}\sum_{i=1}^k \varphi_i(x_i)-c(x^k)$. We therefore have
\begin{align}
    \left|\MOT_{c,\epsilon}(\muk)- \MOT^\epsilon_{c,m,a}(\muk)\right| &\leq (1+e^M/\epsilon)\sum_{i=1}^k \EE_{\mu_i}\left[|\varphi_i^\star-f_i|\right]\nonumber\\
    &\leq (1+e^M/\epsilon)\sum_{i=1}^k \left\|\varphi_i^\star-f_i\right\|_{\infty,\cX_i}\nonumber\\
    &\leq (\epsilon+e^M)k \epsilon^{-1}\max_{i=1,\dots,k}\left\|\varphi_i^\star-f_i\right\|_{\infty,\cX_i}.\label{eq:approx_with_f_bound}
\end{align}
Due to \eqref{eq:approx_with_f_bound}, our goal boils down to bounding the $\sup$-norm error between the optimal dual potentials and the corresponding neural net proxies.
To do that, we are interested in employing the following approximation bound from \cite{sreekumar2021non}, which was previously utilized to bound the approximation error of $f$-divergence neural estimators by a measure of the function smoothness.
To this end, we define the function class
$$
\cC^s_b(\cU)\coloneqq\{f \in \cC^s(\cU): \max_{\alpha: \|\alpha\|_1 \leq s} \|D^\alpha  f \|_{\infty,\cU} \leq  b  \},
$$
where $D^\alpha$, $\alpha=(\alpha_1, \ldots, \alpha_d) \in \ZZ^d_{\geq 0}$, is the partial derivative operator of order $\sum_{i=1}^d\alpha_i.
$
\begin{proposition}[\cite{sreekumar2022neural}, Proposition~10]\label{prop:ne_aprox_smooth}
    Let $\cX\in\RR^d$ be compact and let $g:\cX\to\RR$. Suppose that there exists an open set $\cU\supset\cX$, $b\geq 0$ and $\tilde{g}\in\cC_b^{\skb}(\cU)$ with $\skb\coloneqq\lfloor \frac{d}{2}\rfloor+3$, such that $g=\tilde{g}\big|_{\cX}$. Then, there exist $f\in\cF_{m,d}(\bar{c}_{b,d,\|\cX\|})$ where $\bar{c}_{b,d,\|\cX\|}$ is a constant that depends on $(b,d)$ and is proportional to $\max_{\|\alpha\|_1}\|D^\alpha f\|_{\infty,\cX}$  is given in \cite{sreekumar2022neural},Eqn.~A.15, such that
    $$
    \|f-g\|_\infty\lesssim \bar{c}_{b,d,\|\cX\|} d^{\frac{1}{2}}m^{-\frac{1}{2}}.
    $$
\end{proposition}
A preliminary step to use Proposition \ref{prop:ne_aprox_smooth} is to that for any $i$, $\varphi_i\in\cC_b^{\skb}$, i.e., that the function we aim to approximate adhere to the required smoothness conditions. 
Furthermore, for an explicit characterization of $\bar{c}_{b,d,\|\cX\|}$ in terms of the problem's parameters, the smoothness of the partial derivatives of the dual potentials should also be characterized.
To that end, we propose the following lemma.
\begin{lemma}\label{lemma:ne_smoothness}
    There exist dual EMOT potentials $\left(\varphi_1,\dots,\varphi_k\right)$ for $\MOT_{c,\epsilon}(\muk)$ such that 
    \begin{align}
        &\max_{i=1,\dots,k}\|\varphi_i\|_{\infty,\cX_i}\lesssim_d|\cG_c|\label{eq:smoothness_fn}\\
        &\max_{i=1,\dots,k}\|D^\alpha\varphi_i\|\lesssim_{d,s} 
        (1+\epsilon^{1-s})\left(\mathsf{deg}(\cG_x)\right)^s,\quad 1\leq|\alpha|\leq s.\label{eq:smoothness_der}
    \end{align}
    for any $s\geq 2$ and some constant $C_s$ that depends only on $s$.
\end{lemma}
 The proof of Lemma \ref{lemma:ne_smoothness}, which is given in Supplement \ref{app:ne_smoothness_proof}, is a generalization of \cite{wang2024neural} Lemma 1, which accounts for the dual potential in the bimarginal EOT case. Specifically, given a set of optimal potentials $(\varphi_1^\star,\dots,\varphi_k^\star)$ we can construct a set of dual potentials $(\varphi_1',\dots,\varphi_k')$ that benefit from an explicit representation via the Schr\"odinger system, while agreeing with $(\varphi_1^\star,\dots,\varphi_k^\star)$ $\mu_k$-a.s., rendering them optimal as well for the considered EMOT.
Accordingly, the proof follows from similar arguments.
Using Lemma \ref{lemma:ne_smoothness} we result with the following error on the approximation error
\begin{equation}\label{eq:approx_err}
    \left|\MOT_{c,\epsilon}(\muk)- \MOT^\epsilon_{c,m,a}(\muk)\right|\lesssim_{d,M_{\cX}}\left(1+\epsilon^{-(2+\lfloor \frac{d}{2}\rfloor)}\right)\left(\mathsf{deg}(\cG_c)\right)^{\lfloor \frac{d}{2}\rfloor+3}km^{-\frac{1}{2}}.
\end{equation}


We now turn to bound the estimator error, which is given by the following lemma.
\begin{lemma}[Estimation Error]\label{lemma:est_error}
    Under the considered setting, we have
    \begin{equation}\label{eq:est_err}
        \EE\left[\left| \MOT^\epsilon_{c,m,a}-\widehat{\MOT}_{c,\epsilon}^{m,a}(\muk)\right|\right] \lesssim_{d} n^{-\frac{1}{2}}\max\left\{(1+\epsilon^{-(d+4)})  \left(\mathsf{deg}(\cG_c)\right)^{d+6}, (1+\epsilon^{-(\frac{d}{2} +2)})k\right\}.
    \end{equation}
\end{lemma}
The proof of Lemma \ref{lemma:est_error} is given in Supplement \ref{proof:est_error} and follows the steps of the corresponding estimation error bound from \cite{wang2024neural}. It uses the established smoothness of the integrands to construct bounds on the empirical error using covering and bracketing numbers of the corresponding function classes, combined with maximum inequalities of empirical processes \cite{van1996weak}.
We note that the maximum between the two terms in \eqref{eq:est_err} depends on the cost graph, as $\mathsf{deg}(\cG_c)$ is independent on $k$ for sparse graphs.

Finally, combining \eqref{eq:approx_err} and \eqref{eq:est_err}, we obtain
\begin{align*}
    &\EE\left[\left| \widehat{\MOT}_{c,\epsilon}^{m,a}(\mathbf{X}^{k\times n}) -  \MOT_{c,\epsilon}(\muk)\right|\right]\\
    &\leq (1+\epsilon^{-(\frac{d}{2}+1)})\left(\mathsf{deg}(\cG_c)\right)^{\frac{d}{2}+2}km^{-\frac{1}{2}}+n^{-\frac{1}{2}}\max\left\{ (1+\epsilon^{-(d+4)})\left(\mathsf{deg}(\cG_c)\right)^{d+6} , (1+\epsilon^{-(\frac{d}{2}+2)})k \right\}.
\end{align*}
To obtain a simpler bound, we consider two cases, distinguishing between the dependence of $\mathsf{deg}(\cG_c)$ on $k$. 
In the worst case, when $\mathsf{deg}(\cG_c)=k$, the first term is dominant in the error bound, leaving us with
$$
\EE\left[\left| \widehat{\MOT}_{c,\epsilon}^{m,a}(\mathbf{X}^{k\times n}) -  \MOT_{c,\epsilon}(\muk)\right|\right]\lesssim(1+\epsilon^{-(\frac{d}{2}+1)})k^{d+6}(m^{-\frac{1}{2}}+n^{-\frac{1}{2}}),
$$
while, in the sparse case where $\mathsf{deg}(\cG_c)$ is a constant that is independent of $k$ we have
$$
\EE\left[\left| \widehat{\MOT}_{c,\epsilon}^{m,a}(\mathbf{X}^{k\times n}) -  \MOT_{c,\epsilon}(\muk)\right|\right]\lesssim(1+\epsilon^{-(\frac{d}{2}+1)})k(m^{-\frac{1}{2}}+n^{-\frac{1}{2}}).
$$
This concludes the proof. $\hfill\square$

\begin{remark}[Dependence on dimension]
In the bimarginal case, it is well known that the EOT empirical plug-in estimator attains the parametric rate of convergence $n^{-1/2}$ with a multiplicative factor of $\epsilon^{-d/2}$ \cite{genevay2019sample,mena2019statistical}. Early bounds assumed that both population measures were supported in Euclidean spaces of the same dimension $d$. Recently, \cite{groppe2024lower} showed that if the dimensions of the supports differ, say $d_1\neq d_2$, then $\min(d_1,d_2)$ appears in the empirical convergence rate. This phenomenon is known as the \emph{lower complexity adaptation} principle for EOT, and it also holds for its neural estimator \cite{wang2024neural}.
In contrast, the bound from Theorem~\ref{thm:ne_erro} depends on the largest dimension. The argument used in the bivariate setting achieves a rate depending on the smaller dimension by using the fact that one dual potential determines the other via the $(c,\epsilon)$-transform. Similarly for $k>2$, any one dual potential can~be~expressed in terms of the remaining ones, which reduces the problem from $k$ to $k-1$ potentials. In particular, this allows expressing the largest-dimensional marginal in terms of the others. By analogy to the $k=2$ case, we therefore conjecture that the bound of Theorem~\ref{thm:ne_erro} can be improved from using $\max_i d_i$ to using the \emph{second largest} dimension (which in the $k=2$ case would revert to the minimum). Adjusting our proof accordingly would be an interesting followup.
\end{remark}

\subsection{Proof of Theorem \ref{thm:ne_plan_approx}}\label{proof:ne_plan_approx}
The proof extends the technique from \cite{wang2024neural}.
Let $\mu_1,\dots,\mu_k$ and denote the corresponding random variables $X_1,\dots,X_k$.
We begin by showing that for a pair of couplings $(\pi_\star^\epsilon,\pi_f^\epsilon)$ with $\pi_\star^\epsilon$ being the optimal coupling and optimal potentials $f^\star \coloneqq (f^\star_1,\dots,f^\star_k)$, the KL divergence $\DKL(\pi_\star^\epsilon\|\pi_f^\epsilon)$ is upper bounded by the subtraction of EMOT costs, up to a multiplicative factor.
Specifically,
$$
\DKL(\pi_\star^\epsilon\|\pi_f^\epsilon)\leq\epsilon^{-1}\left(\Gamma(f^\star)-\Gamma(f)\right),
$$
for any continuous $f$.
To that end, we consider the $(c,\epsilon)$-transform, which stems from the Schr\"odinger system of the EMOT \cite{carlier2020differential}, and allows us to represent a single potential in terms of the rest, which for any $f_k$ is given by
\begin{align*}
    \Gamma(f) = \int_{\cX^{k-1}}\bar{f}d\bar{\mu}+\int_{\cX_k}f^{c,\epsilon}d\mu_k,\quad f^{c,\epsilon}(y)\coloneqq-\epsilon\log\left(\int_{\cX^{k-1}}\exp\left(\frac{\bar{f}(\bar{x},y)}{\epsilon}\right)d\bar{\pi}(\bar{x})\right),
\end{align*}
where $\bar{x}=(x_1,\dots,x_{k-1})$, $\bar{f}(\bar{x})=\sum_{i=1}^{k-1}f_i(x_i)$, $\bar{\mu}=(\mu_1,\dots,\mu_{k-1})$ and $\bar{\pi}=\int_{\cX}\pi(\bar{x},y)d\mu_k(y)$.
We define the energy function $E_{\bar{\pi}}$.
Furthermore, note that in this setting, the conditional plan is given by:
$$
\frac{d\pi(\bar{x}|y)}{d\bar{x}}=\frac{\exp\left(\frac{  \tilde{f}(\bar{x}) - c(\bar{x},y)}{\epsilon}\right)}{\int \exp\left( \frac{ \tilde{f}(\bar{x}') - c(\bar{x}',y)}{\epsilon}\right)d\bar{x}}.
$$
Define $F_{\bar{f}}(y) \coloneqq \int_{\cX^{k-1}}\exp\left(\frac{\bar{f}(\bar{x})-c(\bar{x},y)}{\epsilon}\right)d\bar{x}$ and $Z\coloneqq \int_{\cX^{k-1}}\exp\left(-E_{\bar{\pi}}(\bar{x})\right)$.
We therefore have $\bar{f}^{c,\epsilon}(y) = -\epsilon \log (F_{\bar{f}}(y))+\epsilon\log(Z)$.
We have the following:

\begin{align*}
&\Gamma(f_\star)- \Gamma(f)\\
&= \int_{\cX^k} c\,d\pi_\star^\epsilon +\epsilon\DKL(\pi_\star^\epsilon\|\otimes_{i=1}^k\mu_i) -\int_{\cX^{k-1}} \bar{f}\,d\bar{\pi} -\int_{\cX_k} f^{c,\epsilon}\,d\mu_k\\
&\stackrel{a}{=} \int_{\cX^k} \bigl(c(\bar{x},y)-\tilde{f}(\bar{x})\bigr)\,d\pi_\star^\epsilon(\bar{x},y)
-\int_{\cX_k}h(\cdot|y)d\mu_k(y) + h(\bar{\pi}) 
+\epsilon \int_{\cX_k} \log\bigl(F_{\tilde{f}}(y)\bigr)\,d\mu_k(y)+ \DKL\left(\bar{\pi}\|\otimes_{i=1}^{k-1}\mu_i\right)\\
&= -\,\epsilon \int_{\cX^k} \frac{\tilde{f}(x)-c(\bar{x},y)}{\epsilon}\,d\pi_\star^\epsilon(x,y)
+\epsilon \int_{\cX^k} \log\bigl(F_{\tilde{f}}(y)\bigr)\,d\pi_\star^\epsilon(x,y)
-\epsilon \int_{\cX_k} h\bigl(\pi_\star^\epsilon(\cdot\mid y)\bigr)\,d\mu_k(y)+ \DKL\left(\bar{\pi}\|\otimes_{i=1}^{k-1}\mu_i\right)\\
&= -\,\epsilon \int_{\cX^k}
\log\!\Bigl(\tfrac{1}{F_{\tilde{f}}(y)} \,\exp\!\bigl(\tfrac{\tilde{f}(x)-c(\bar{x},y)}{\epsilon}\bigr)\Bigr)
\,d\pi_\star^\epsilon(x,y)
-\epsilon \int_{\cX_k} h\bigl(\pi_\star^\epsilon(\cdot\mid y)\bigr)\,d\mu_k(y)+ \DKL\left(\bar{\pi}\|\otimes_{i=1}^{k-1}\mu_i\right)\\
&= -\,\epsilon \int_{\cX^k}
\log\!\Bigl(\tfrac{d\pi_f^\epsilon(x\mid y)}{d\pi_\star^\epsilon(x\mid y)}\Bigr)
\,d\pi_\star^\epsilon(x,y)
-\epsilon \int_{\cX_k} h\bigl(\pi_\star^\epsilon(\cdot\mid y)\bigr)\,d\mu_k(y)+ \DKL\left(\bar{\pi}\|\otimes_{i=1}^{k-1}\mu_i\right)\\
&= -\,\epsilon \int_{\cX_k}\!\int_{\cX^{k-1}}
\log\!\Bigl(\tfrac{d\pi_f^\epsilon(x\mid y)}{d\pi_\star^\epsilon(x\mid y)}\Bigr)\,
d\pi_\star^\epsilon(x\mid y)\,d\mu_k(y)+ \DKL\left(\bar{\pi}\|\otimes_{i=1}^{k-1}\mu_i\right)\\
&= \epsilon \int_{\cX_k}\!\int_{\cX^{k-1}}
\log\!\Bigl(\tfrac{d\pi_\star^\epsilon(x\mid y)}{d\pi_f^\epsilon(x\mid y)}\Bigr)\,
d\pi_\star^\epsilon(x\mid y)\,d\mu_k(y)+ \DKL\left(\bar{\pi}\|\otimes_{i=1}^{k-1}\mu_i\right)\\
&= \epsilon \int_{\cX^k}
\log\!\Bigl(\tfrac{d\pi_\star^\epsilon(x,y)}{d\pi_f^\epsilon(x,y)}\Bigr)\,
d\pi_\star^\epsilon(x,y)+ \DKL\left(\bar{\pi}\|\otimes_{i=1}^{k-1}\mu_i\right)\\
&=\epsilon\DKL\!\bigl(\pi_\star^\epsilon \,\big\|\,\pi_f^\epsilon\bigr)+ \DKL\left(\bar{\pi}\|\otimes_{i=1}^{k-1}\mu_i\right),
\end{align*}
where $h(\mu) = -\int_\cX\log\left(\frac{d\mu}{dx}\right)d\mu(x)$ is the differential entropy of $\mu$ and (a) follows from
\begin{align*}
    \DKL(\pi_\star^\epsilon\|\otimes_{i=1}^k\mu_i) &= \EE\left[\log\frac{d\pi_\star^\epsilon}{d(\otimes_{i=1}^k\mu_i)}\right]\\
    &= \EE\left[\log\frac{d\pi_\star^\epsilon}{d\mu_k}\right]- \EE\left[\log\bar{\pi}_\star^\epsilon\right] + \EE\left[\log\frac{d\bar{\pi}_\star^\epsilon}{d(\otimes_{i=1}^{k-1}\mu_i)}\right]\\
    &= -\int_{\cX_k}h(\cdot|y)d\mu_k(y) + h(\bar{\pi}) + \DKL\left(\bar{\pi}\|\otimes_{i=1}^{k-1}\mu_i\right).
\end{align*}
Next, we bound the EMOT cost difference.
We have the following:
\begin{align*}
    &\Gamma(f_\star)- \Gamma(f) \\
    &\leq \underbrace{\MOT_{c,\epsilon}(\muk)-\MOT_{c,\epsilon}^{m,a}(\muk)}_{(i)} + \underbrace{\MOT_{c,\epsilon}^{m,a}(\muk) - \widehat{\MOT}_{c,\epsilon}(\textbf{X}^{n,k})}_{(ii)} + \underbrace{\hat{\Gamma}(\hat{f}^\star)-\Gamma(\hat{f}^\star)}_{(iii)},
\end{align*}
where $\Gamma(\hat{f}^\star)$ is the EMOT cost, calculated from the optimal neural net within the class $\cF_{\mathsf{nn}}^{\mspace{1mu}m}(a)$.
Note that (i) and (ii) quantify the NEMOT approximation and estimation error, respectively. Therefore, we can simply use the bounds from Supplement \ref{proof:ne_error}.
Furthermore, for (iii), as noted in \cite{wang2024neural}, we have
\begin{align*}
    \EE\left[\hat{\Gamma}(\hat{f}^\star)-\Gamma(\hat{f}^\star)\right]&\leq\EE\left[\sup_{f\in\cF_{\mathsf{nn}}^{\mspace{1mu}m}(a)}\left|\hat{\Gamma}(\hat{f}^\star)-\Gamma(\hat{f}^\star)\right|\right]\\
    &\leq \sum_{j=1}^k\frac{1}{\sqrt{n}}\EE\left[\sup_{f\in\cF_{\mathsf{nn}}^{\mspace{1mu}m}(a)}\frac{1}{\sqrt{n}}\left|\sum_{i=1}^n(f_j(X_{ij})-\EE_{\mu_j}[f_j])\right|\right],
\end{align*}
which can correspondingly bounded by the estimation error.
Combining the bounds, we result with the same bound as the neural estimation bound \eqref{eq:nemot_est_bound}. This concludes the proof. $\hfill\square$

\subsection{Proof of Lemma \ref{lemma:ne_smoothness}}\label{app:ne_smoothness_proof}
Fix $i\in[1,\dots,k]$. 
We start with the bound of $\|\varphi^\star_i\|_{\infty,\cX_i}$.
By the invariance of the EMOT to additive constants we may without loss of generality take optimal MOT potentials such that $\int_{\cX_i} \varphi_id\mu_i = \frac{1}{k}\MOT_{c,\epsilon}$.
We consider a set of dual potentials $(\varphi_i)_{i=1}^k$ that satisfy the Schr\"odinger system with the optimal potentials, i.e.,
$$
\int\exp\left(\sum_{i=1}^k \varphi_i^\star(x_i)-c(x^k)\right)d\mu^{-i}(x^{-i}) = 1,\quad \mu_i-\text{a.e.},
$$
where $\cX^{-i} 
(\times_{j=1}^k \cX_j)\setminus \cX_i$ and similarly $\mu^{-i}\coloneqq \otimes_{j\in[1,\dots,k]\setminus i}\mu_j$.
Thus, for any $i\in[1,\dots,k]$ we have
\begin{equation}\label{eq:schrodinger}
    \varphi_i(x) = -\epsilon\log\int_{\cX^{-i}}\exp\left(\frac{\sum_{-j}\varphi_j(x_j)-c(x^k)}{\epsilon} d\mu^{-i}(x^{-i})\right).
\end{equation}
To obtain the desired bound, we derive a uniform bound over $\cX_i$. First, by Jensen's inequality, we have
\begin{align*}
    \varphi_i(x_i) &\leq \int_{\cX^-i}c(x^k) - \sum_{j=1,j\neq i}^k \varphi_j^\star(x_j) d\mu^{-i}(x^{-i})\\
    &\leq |\cG_c|C(d) -\frac{k-1}{k}\MOT_{c,\epsilon}(\muk)\\
    &\leq |\cG_c|C(d),
\end{align*}
where the second inequality follows from the choice $\int_{\cX_i} \varphi_i^\star d\mu_i = \frac{1}{k}\MOT_{c,\epsilon}(\muk)\geq 0$ and the decomposition of $c(x^k)$ into its pairwise components, each of which upper bounded $C(d_i,d_j)$, which we uniformly upper bound with $C(d)$ such that $d=\max_{i=1}^k d_i$.
For example, when the pairwise cost is quadratic, $C(d_i,d_j)=2\max(d_i,d_j)$.
Next, we have
\begin{align*}
    \varphi_i(x_i) &\leq \epsilon\log\int_{\cX^-i}\exp\left(\frac{c(x^k) - \sum_{j=1,j\neq i}^k \varphi_j^\star(x_j) }{\epsilon}\right)d\mu^{-i}(x^{-i})\\
    &\leq \epsilon\log\int_{\cX^-i}\exp\left(\frac{(1 + \frac{k-1}{k})|\cG_c|C(d) }{\epsilon}\right)d\mu^{-i}(x^{-i})\\
    &\leq 2C(d)|\cG_c|.
\end{align*}
Here, the second inequality follows from $\MOT_{c,\epsilon}(\muk) \leq |\cG_c|C(d)$.
As the bound is uniform we can conclude that
$$
\max_{i=1,\dots,k}\|\varphi_i\|_{\infty,\cX_i}\lesssim_d|\cG_c|.
$$
Next, note that $(\phi_1,\dots,\phi_k)$ are also optimal potentials of the EMOT, as by Jensen's inequality we have
\begin{align*}
    \sum_{i=1}^k (\varphi_i^\star-\varphi_i)d\mu_i &\leq \sum_{i=1}^k \epsilon\log\int_{\cX_i}\exp\left(\frac{\varphi_i^\star-\varphi_i}{\epsilon}\right)d\mu_i\\
    &=\sum_{i=1}^k\epsilon\log\int_{\cX^k}\exp\left(\frac{\sum_{j=1}^k\varphi_j^\star-c}{\epsilon}\right)d(\otimes_{j=1}^k\mu_j)\\
    &=0.
\end{align*}
By the concavity of the logarithm function we can conclude that $\varphi_i=\varphi^\star_i$ $\mu_i$-a.s. for $i=1,\dots,k$.

Next we bound the magnitude of the partial derivative. The differentiability of the dual potentials is granted from their definition. To bound the partial derivative we will use the Faa di Bruno Formula (\cite{constantine1996multivariate} Corollary 2.10), which for a multi-index $\alpha$, provides us with the following characterization of the partial derivative of $\varphi$ w.r.t. $\alpha$ as follows
\begin{align}
    &-D^\alpha(\varphi_i)(x) =\epsilon\sum_{r=1}^\alpha\sum_{p(\alpha,r)}\frac{\alpha!(r-1)!(-1)^{r-1}}{\prod_{j=1}^{|\alpha|}(k_j!)(\beta_j!)^{k_j}}\nonumber\\
    &\hspace{5cm}\times\prod_{j=1}^{|\alpha|}\left( \frac{D^{\beta_j}\int\exp\left( \frac{\sum_{\ell=1,\ell\neq i}^k\varphi^\star_\ell(x_k)-c(x^k) }{\epsilon}d\mu^{-i}(x^{-i}) \right)}{\int\exp\left( \frac{\sum_{\ell=1,\ell\neq i}^k\varphi^\star_\ell(x_k)-c(x^k) }{\epsilon}d\mu^{-i}(x^{-i}) \right)} \right),\label{eq:faadibruno}
\end{align}
where $p(\alpha,r)$ is the collection of all tuples $(k_1,\dots,k_{|\alpha|;\beta_1,\dots,\beta_|\alpha|})\in\NN^{|\alpha|}\times\NN^{d_i\times\alpha}$ satisfying $\sum_{i=1}^{|\alpha|}k_i=r,\sum_{i=1}^{|\alpha|}k_i\beta_i=\alpha$ and for which there exit $s\in(1,\dots,|\alpha|)$ such that $k_i=0$ and $\beta_i=0$ for all $i\in(1,\dots,|\alpha|-s)$, $k_i>0$ for all $i\in(|\alpha|-s+1,\dots,|\alpha|)$ and $0\prec \beta_{|\alpha|-s+1}\prec\dots\prec \beta_{|\alpha|}$. We refer the reader to \cite{constantine1996multivariate} for more information on the Faa di Bruno formula.
For our purpose, it is sufficient to bound $D^{\beta_j}\int\exp\left( \sum_{\ell=1,\ell\neq i}^k\varphi^\star_\ell(x_k)-c(x^k)\right) d\mu^{-i}(x^{-i})$.
To that end, we apply the Faa di Bronu formula to $D^{\beta_j}\exp(-c(x^k)/\epsilon)$ yielding
$$
D^{\beta_j}\exp(-c(x^k)/\epsilon)\sum_{r'=1}^{|\eta_j|}\left(\frac{-1}{2\epsilon}\right)^{r'}\sum_{p(\beta_j,r')}l(\beta_j,r',\mathbf{k},\mathbf{\eta})\exp(-c(x^k)\prod_{\ell=1}^{\beta_j}D^{\eta_\ell}c(x^k),
$$
and the cost derivative can be bounded using the pairwise cost bound, i.e., for any $\eta_\ell$
$$
D^{\eta_\ell}c(x^k) = \sum_{j\in N(X_i)}D^{\eta_\ell}c(x_i,x_j) \leq C'(d)\mathsf{deg}(\cG_c).
$$
Here $N(X_i)$ are the neighbors of the variable $X_i$ in the cost graph $\cG_c$, $C'(d)$ is a uniform bound on the pairwise costs derivatives that depends only on $d$, and $\mathsf{deg}(\cG_c)$ is the degree of the graph $\cG_c$. \
The sum $\sum_{r'=1}^{|\eta_j|}\left(\frac{-1}{2\epsilon}\right)^{r'}$ is upper bounded by $\epsilon^{-|\beta_j|}$ when $0\leq\epsilon\leq 1$, and by $\epsilon^{-1}$ when $\epsilon>1$. Consequently, we have uniformly on $\cX_i$
$$
|D^\alpha\varphi_i(x_i)| \leq C'(|\alpha|,d)(1+\epsilon^{1-|\alpha|})\left(\mathsf{deg}(\cG_c)\right)^{|\alpha|},
$$
which concludes the proof.$\hfill\square$

\subsection{Proof of Lemma \ref{lemma:est_error}}\label{proof:est_error}
We start by decomposing the expression at hand. To that end we define $\cF_{k,m}(a) = \bigcup_{i=1}^k\cF_{m,d_i}(a)$. We have the following
\begin{align*}
    &\EE\left[\left| \MOT^\epsilon_{c,m,a}-\widehat{\MOT}^\epsilon_{a,m}(\muk)\right|\right] \\
    &\leq n^{-\frac{1}{2}}\sum_{i=1}^k\left(\EE\left[\sup_{f_i\in\cF_{m,d_i}(a)} \left| n^{-\frac{1}{2}}\sum_{j=1}^n f_i(x_{ij})-\EE_{\mu_i}\left[ f_i \right] \right| \right]\right)\\
    &+ n^{-\frac{1}{2}}\Bigg(\EE\Bigg[\sup_{(f_1,\dots,f_k)\in\cF_{k,m}(a)} \Bigg| n^{-\frac{1}{2}}\sum_{j=1}^n 
    e^{\left( \sum_{i=1}^k f_i(x_{ij}) - c(\bxk_j) \right)/\epsilon}
    -\EE\Bigg[ e^{\left( \bigoplus_{i=1}^k f_i - c \right)/\epsilon} \Bigg] \Bigg| \Bigg]\Bigg).
\end{align*}
Te first term contains $k$ distinct error terms for each of the neural nets $f_i$, $i\leq k$.
This bound is therefore effectively reduced to the one considered in \cite{wang2024neural}. Consequently, we can repeat the arguments in \cite{wang2024neural} Eqn.~24, which bound the expected error using maximal inequalities of empirical processes, combined with a bound on the covering number of the dual potential function class. Thus, for each $i\leq k$ we have
$$
\EE\left[\sup_{f_i\in\cF_{m,d_i}(a)} \left| n^{-\frac{1}{2}}\sum_{j=1}^n f_i(x_{ij})-\EE_{\mu_i}\left[ f_i \right] \right| \right] \leq ad_i^{3/2}.
$$
To bound the second term we need to account for the smoothness of the exponential term we aim to estimate, which was already established in Supplement \ref{app:ne_smoothness_proof}.
We follow the step in \cite{wang2024neural}, with an adjustment to the exponential term smoothness bound. To avoid heavy notation, we denote $\sF(\bxk)\coloneqq e^{\left( \sum_{i=1}^k f_i(x_{ij}) - c(\bxk_j) \right)/\epsilon}$. We have
\begin{align*}
    \EE\left[\sup_{(\Vec{f}_{\theta,k})\in\cF_{k,m}(a)} \left| n^{-\frac{1}{2}}\sum_{j=1}^n 
    \sF(\bxk_j)
    -\EE\left[ \sF\right] \right| \right] 
    &\stackrel{(a)}{\lesssim} \EE\left[ \int_{0}^\infty \sqrt{\log N(\delta,\cF_{k,m}(a),\|\cdot\|_{2,\otimes_{i=1}^k(\mu_i)_n}) }d\delta \right]\\
    &\leq \int_{0}^\infty \sqrt{\sup_{\gamma\in\cP(\cxk)}\log N(\delta,\cF_{k,m}(a),\|\cdot\|_{2,\gamma}) }d\delta \\
    &\lesssim\sup_{\gamma\in\cP(\cxk)}\int_{0}^{12a} \sqrt{\log N(\delta,\cF_{k,m}(a),\|\cdot\|_{2,\gamma}) }d\delta\\
    &\stackrel{(b)}{\lesssim} \sup_{\gamma\in\cP(\cxk)}\int_{0}^{12a} \sqrt{\log N_{[]}(\delta,\cF_{k,m}(a),\|\cdot\|_{2,\gamma}) }d\delta\\
    &\stackrel{(c)}{\lesssim} K_s\int_{0}^{12a}\left( \frac{C'(s,d)(1+\epsilon^{1-s})\left(\mathsf{deg}(\cG_c)\right)^{s}}{2\delta} \right)^{\frac{d}{2s}}d\delta\\
    &\lesssim \tilde{C}(s,d) a (1+\epsilon^{1-s})\left(\mathsf{deg}(\cG_c)\right)^{s}
\end{align*}
where $\Vec{f}_{\theta,k}\coloneqq(f_{\theta_1},\dots,f_{\theta_k})$ $(a)$ is an upper bound in terms of the function class covering number, which follows from \cite{van1996weak} Corollary~2.8.2, since $n^{-\frac{1}{2}}\sum_{i=1}^n \sigma_j \sF(\bxk_j)$ is Sub-Gaussian w.r.t. the pseudo metric $\|\cdot\|_{2,\otimes_{i=1}^k(\mu_i)_n}$ whenever $(\sigma_i)_{i=1}^n$ are Rademacher random variables, $(b)$ upper bounds the covering number with the bracketing number and $(c)$ utilizes \cite{van1996weak} Corollary~2.7.2 which upper bounds the bracketing number in terms of the smoothness parameter of the function class with a constant $K_s$ that depends only in $s$ and $C(s,d)$ is the constant from Supplement \ref{app:ne_smoothness_proof}.

Finally, by setting $a=\bar{c}_{b,d}$ and combining both bounds, we have
$$
\EE\left[\left| \MOT^\epsilon_{c,m,a}-\widehat{\MOT}^\epsilon_{a,m}(\muk)\right|\right] \lesssim_{d} n^{-\frac{1}{2}}\max\left\{(1+\epsilon^{-(d+4)})  \left(\mathsf{deg}(\cG_c)\right)^{d+6}, (1+\epsilon^{-(\frac{d}{2} +2)})k\right\}.
$$
This concludes the proof. $\hfill\square$

\subsection{Proof of Proposition \ref{prop:mgw_cost_approx}}\label{proof:mgw_cost_approx}

Let $\pi_0$ be the optimal alignment plan of the unregularized MGW problem. The existence of $\pi_0$ is guaranteed as it is the solution of a linear program.
Let $\ell\in(0,1]$. For any $d\geq1$ and $k\in\ZZ$ let the $l$-sided Euclidean cube be
$$
Q^\ell_u\coloneqq \prod_{i=1}^d \left[ u_m\ell,(u_m+1)\ell \right)\subset\RR^d,\qquad Q^\ell_{\ui}=\times_{j=1}^k Q^\ell_{i_j}, 
$$
where $\ui\coloneqq(i_1,\dots,i_k)$ is a $k$ multi-index. Define the block approximation of $\pi_0$ within $Q^\ell_{\ui}$ as
$$
\pi^\ell\Big|_{Q^\ell_{\ui}} = \frac{\pi_0(Q^\ell_{\ui})}{\prod_{j=1}^k \mu_j (Q^\ell_{i_j})}\otimes_{j=1}^k \mu_j\Big|_{Q^\ell_{i_j}}.
$$
We have
\begin{equation}
    0 \leq \MGW^\epsilon(\mu^k) - \MGW^2(\mu^k) \leq G(\pi^\ell) - G(\pi^\ell) +\epsilon\DKL(\pi^\ell\|\otimes_{i=1}^k\mu_i).
\end{equation}
where 
$$
G(\pi) = \int \Delta(\bxk,\byk)  d\pi\otimes\pi(\bxk,\byk)
$$
The first inequality from MGW attaining a lower value than the entropic counterpart and the nonlinearity of the KL divergence. The second inequality follows as inserting $\pi^\ell$ into $G(\cdot)$ can only increase its value compared to the optimal solution.

We begin by bounding the cost terms. We denote with $\cX^k$ the product of spaces. We have
\begin{align*}
    G(\pi^\ell)-G(\pi^0) &= \int_{\cX^k}\sum_{(i,j)\in\cE}\left| \|x_i-y_i\|^2 - \|x_j-y_j\|^2 \right|^2d\pi^\ell\otimes\pi^\ell(\bxk,\byk)\\
    &\qquad\qquad-\int_{\cX^k}\sum_{(i,j)\in\cE}\left| \|x_i-y_i\|^2 - \|x_j-y_j\|^2 \right|^2d\pi^0\otimes\pi^0(\bxk,\byk)\\
    &= \sum_{(i,j)\in\cE}\Bigg( \int_{\cX^k}\left| \|x_i-y_i\|^2 - \|x_j-y_j\|^2 \right|^2d\pi^\ell\otimes\pi^\ell(\bxk,\byk)\\
    &\qquad\qquad\qquad - \int_{\cX^k}\left| \|x_i-y_i\|^2 - \|x_j-y_j\|^2 \right|^2d\pi^\ell\otimes\pi^\ell(\bxk,\byk) \Bigg).
\end{align*}
Next, marginalize each integral and cancel out terms that only depend on marginal distributions to result with
\begin{align}
    G(\pi^\ell)-G(\pi^0) &\lesssim \sum_{(i,j)\in\cE}\int_{\cX_i\times\cX_j}\|x_i\|^2\|x_j\|^2d(\pi^\ell-\pi^0)(x_i,x_j) \nonumber\\
    &\qquad+ \sum_{\substack{1\leq m \leq d_i\\1\leq p \leq d_j}}\left| \left(x_{im}x_{jp}d\pi^\ell(x_i,x_j)\right)^2 - \left(x_{im}x_{jp}d\pi^0(x_i,x_j)\right)^2 \right|\label{eq:mgw_bound_ref_zhang}.
\end{align}
To bound the pairwise terms in \eqref{eq:mgw_bound_ref_zhang} we utilize the result from \cite{zhang2022gromov}, 
to result with
\begin{equation}\label{eq:mgw_proof_bounded_g}
    G(\pi^\ell)-G(\pi^0) \lesssim \ell \sum_{(i,j)\in\cE}\left(\sqrt{d_i}+\sqrt{d_j}\right)^6\left(1+M_4(\mu_i)+M_4(\mu_j)\right),
\end{equation}
where $M_4(\mu)$ is the fourth moment of $\mu$.
Next, we result the KL term. First, we upper bound the KL term as follows
\begin{align*}
    \DKL(\pi^\ell\|\otimes_{i=1}^k\mu_i) &= \sum_{\ui\in\ZZ^{\sum_{i=1}^k d_i}}\log\left(\frac{\pi^0(Q^\ell_{\ui})}{\prod_{j=1}^k \mu_{j}(Q^\ell_{i_j})}\right)\pi^0(Q^\ell_{\ui})\\
    &=\sum_{\ui\in\ZZ^{\sum_{i=1}^k d_i}}\log\left(\pi^0(Q^\ell_{\ui})\right)\pi^0(Q^\ell_{\ui}) - \sum_{j=1}^k\sum_{i_j\in\ZZ^{d_i}}\log\left(\mu_j(Q^\ell_{i_j})\right)\mu_j(Q^\ell_{i_j})\\
    &\leq  - \sum_{j=1}^k\sum_{i_j\in\ZZ^{d_i}}\log\left(\mu_j(Q^\ell_{i_j})\right)\mu_j(Q^\ell_{i_j})
\end{align*}
To upper bound each of $k$ entropic elements we leverage the tools developed in \cite{zhang2022gromov}. First, we obtain a differential entropy term using a piecewise constant measure. then, we bound this differential entropy with a Gaussian entropy with the same covariance matrix. Finally, we construct an entry-wise bound on the resulting covariance matrix, which leaves us with a bound of the form
\begin{align}
    \DKL(\pi^\ell\|\otimes_{i=1}^k\mu_i) \leq \sum_{i=1}^k d_i+d_i\log(d_i !(M_2(\mu_i)+d_i(1+M_1(\mu_i)))) - d_i\log\ell\label{eq:mgw_proof_dkl_bound}.
\end{align}
We denote the terms in \eqref{eq:mgw_proof_dkl_bound} that do not depend on $\ell$ with $(C_{d,i})_{i=1}^k$.
Combining \eqref{eq:mgw_proof_bounded_g} and \eqref{eq:mgw_proof_dkl_bound}, we result with the bound
\begin{equation}
    \MGW^\epsilon(\mu^k) - \MGW^2(\mu^k) \lesssim \ell \sum_{(i,j)\in\cE}\left(\sqrt{d_i}+\sqrt{d_j}\right)^6\left(1+M_4(\mu_i)+M_4(\mu_j)\right) + \epsilon\sum_{i=1}^k \left(C_{d,i}-d_i\log\ell\right).
\end{equation}
We seek to optimize the above bound w.r.t. $\ell$. Taking the derivative w.r.t. $\ell$, we have
$$
\ell^\star = \frac{\epsilon \sum_{i=1}^k d_i}{\sum_{(i,j)\in\cE}\left(\sqrt{d_i}+\sqrt{d_j}\right)^6\left(1+M_4(\mu_i)+M_4(\mu_j)\right)}.
$$
Take $l = \max(l^\star,1)$. Assuming $k$ is a fixed given parameter of the system, we have
$$
\MGW^\epsilon(\mu^k) - \MGW^2(\mu^k) \lesssim_{k,(d_i)_{i=1}^k}\epsilon\log\left(\frac{1}{\epsilon}\right).
$$
Next, we simplify the bound to result with an explicit dependence on $k$ and the corresponding cost graph $\cG_c$.
Taking $M=\max_{i=1}^k M_4(\mu_i)$, $d=\max_{i=1}^k d_i$, we have
$$
\MGW^\epsilon(\mu^k) - \MGW^2(\mu^k) \lesssim \epsilon k d + \epsilon k C_d - \epsilon k d \log\left(\epsilon k d\right) + \epsilon k d \log\left(64 d^3 (1+2M) \mathsf{deg}(\cE)\right),
$$
where $\mathsf{deg}(V)$ is the degree of the graph $V$.
Therefore, our bound can be represented in terms of both $\epsilon$ and $k$ as 
$$
\MGW^\epsilon(\mu^k) - \MGW^2(\mu^k) \lesssim_d \epsilon k \left(\log \frac{\mathsf{deg}(\cG_c)}{k\epsilon}\right).
$$
This concludes the proof. $\hfill\square$

\subsection{Proof Of Proposition \ref{prop:mgw_s2}}\label{appendix:emgw_lin}
For simplicity of presentation, we consider a circle cost graph. The proof readily generalizes to any cost that decomposes into pairwise terms.
Under the circular cost, we optimize $k$ alignment matrices $\rA_1,\dots,\rA_{k}$ and the plan-dependent term is given by
$$
\sS^2_\epsilon = \inf_\pi\sum_{i=1}^k-4\int \|x_{i}\|^2\|x_{i+1}\|^2d\pi(x_{i},x_{i+1})-8\int\langle x_{i},y_{i} \rangle \langle x_{i+1},y_{i+1} \rangle d\pi\otimes\pi(x_{i},x_{j},y_{i},y_{j})+\epsilon\DKL(\pi\|\muk),
$$
where, recall we use the convention $i=i\mod k$.
Our goal is to eliminate the quadratic dependence on $\pi$.
First note, that each term in the sum considers a specific marginalized term of the plan $\pi$ w.r.t. the coordinates $(i,i+1)$.
Focusing on a specific pair, e.g. $(i,i+1)=(1,2)$, denote the resulting term with $\sS(1,2)$.
Denote with $d_i$ the dimension of the $i$th Euclidean space and recall that $M_{ij}=\sqrt{M_2(\mu_i)M_2(\mu_j)}$. We do that by inserting auxiliary matrices to linearize the quadratic term. We have the following:
\begin{align*}
    \sS_{1}&=-4\left(\int \|x_{1}\|^2\|x_{2}\|^2d\pi(x_{1},x_{2})-8\int\langle x_{1},y_{1} \rangle \langle x_{2},y_{2} \rangle d\pi\otimes\pi(x_{1},x_{2},y_{1},y_{2})\right)\\
    &= -4\int \|x_{1}\|^2\|x_{2}\|^2d\pi(x_{1},x_{2})+\sum_{\substack{1\leq\ell_1\leq d_1 \\ 1\leq\ell_2\leq d_2}} \inf_{|a_{\ell_1,\ell_2}|\leq\frac{M_{1,2}}{2}}32\left( a_{\ell1_,\ell_2}^2 - \int a_{\ell1_,\ell_2}x_{1,\ell_1}x_{2,\ell_2}d\pi(x_{1,\ell_1}x_{2,\ell_2})\right)\\
    &= \inf_{\rA_{1,2}\in\cD_{M_{1,2}}}-4\int \|x_{1}\|^2\|x_{2}\|^2d\pi(x_{1},x_{2})+8\sum_{\substack{1\leq\ell_1\leq d_1 \\ 1\leq\ell_2\leq d_2}} 32\left( a_{\ell1_,\ell_2}^2 - \int a_{\ell1_,\ell_2}x_{1,\ell_1}x_{2,\ell_2}d\pi(x_{1,\ell_1}x_{2,\ell_2})\right)\\
    &= \inf_{\rA_{1}\in\cD_{M_{1,2}}}32\|\rA\|^2_\mathrm{F}-4\int \|x_{1}\|^2\|x_{2}\|^2d\pi(x_{1},x_{2})+8\sum_{\substack{1\leq\ell_1\leq d_1 \\ 1\leq\ell_2\leq d_2}} 32\left( - \int a_{\ell1_,\ell_2}x_{1,\ell_1}x_{2,\ell_2}d\pi(x_{1,\ell_1}x_{2,\ell_2})\right),
\end{align*}
where we introduced $a_{\ell_1,\ell_2}$ whose optimum is achieved at $\frac{1}{2}x_ix_{i+1}d\pi(x_i,x_{i+1})$, which means we can restrict the optimization of $M_{i,i+1}$.
The same strategy can be applied to each term in the sum, thereby introducing $k$ auxiliary alignment matrices $\rA_{1},\dots,\rA_{k}$, with $\rA_i\in\RR^{d_1,d_{i+1}}\cap M_{1,1_{i+1}}$.
Finally, arrange the matrices in a block matrix $\rA$ with $k$ blocks, this having $\|\rA\|^2_\mathrm{F}=\sum_{i=1}^k\|\rA_i\|^2_\mathrm{F}$. Due to lower semicontinuity of the expression in $\pi$ and $\rA$, we have
\begin{align}
    \sS^2_\epsilon = \sup_{\rA\in\cD_M} \inf_{\pi \in \Pi(\muk)} 32\|\rA\|^2_\mathrm{F} + \int_{\cX^k} c_\rA(\bxk)d\pi(\bxk) + \epsilon\DKL(\pi\|\muk).
\end{align}
This concludes the proof. $\hfill\square$


\section{Additional Implementation Details}\label{app:implementation}
All NEMOT models are implemented in PyTorch \cite{paszke2019pytorch}. 
The models are trained with the Adam optimizer \cite{kingma2014adam} with an initial learning rate of $5\times 10^{-5}$.
As the neural estimation batch loss \eqref{eq:ne_emot} considers an exponential term, we introduce a gradient norm clipping with the norm clipping parameter of $0.1$. We consider an exponential learning rate decay with decay value of $0.5$ and scheduling rate of $5$ epochs.
Unless stated otherwise, we consider a batch size of $b=64$ and train the network for 50 epochs to ensure convergence of the loss for small values of $n$.
However, in all training sessions the estimate saturated after less than 20 epochs.
The NEMOT networks are implemented with simple feedforward MLPs. The networks consist of 3 layers with output sizes $[10K, 10K, 1]$ with $K\coloneqq\min(10d,80)$, where $d$ is the data dimension.
Each layer output is followed with a ReLU activation.
When calculating the neural plan, we apply normalization by the sum of its entries to ensure it is a proper joint distribution over the product space.
In practice, we found that drawing gradients for all the $k$ networks improved the NEMOT convergence, while at a negligible cost of runtime. This method of course introduces a trade-off, which is made crucial in the large $k$ regime.

\paragraph{GMM data.}
The GMM dataset is constructed as a dataset of $k$ i.i.d. GMM variables.
Each GMM is a mixture of $m=3$ Gaussian variables with different means which are generated in $[-1,1]^{d/2}$ and are mapped into $\RR^d$ using an orthonormal matrix.
All Gaussian components have standard deviation $\sigma=0.1$.
The GMM variable is then obtained by randomly sampling the Gaussian variables.

\section{Computational Complexity of NEMOT via U-statistics}\label{appendix:u-stat}
Herein, we consider the U-statistic estimator of the exponential term in the NEMOT.
For a given dataset $\mathbf{X^{n,k}}$, the $k$-sample U-statistic estimator considers all possible $k$-tuples that can be generated from the dataset and takes the sample mean over the exponential term evaluated over these tuples.
Formally it is given by the following expression
$$
\hat{E}_\sU \coloneqq\frac{1}{n^k}\sum_{j_1,\dots,j_k}e^{\left(\sum_{i=1}^k f_{\theta_i}(X_{j_i}) -c(X_{j_1},\dots,X_{j_k})\right)/\epsilon}+\epsilon.
$$
where $(j_1,\dots,j_k)$ are all possible $k$-tuples, in which each element takes a value in $\{1,\dots,n\}$.
Generally $U$-statistics form low variance estimators.
However, for a given batch size $b$, to calculate $\hat{E}_\sU$, one must take the average of a data structure with $b^k$.
This is indeed a significant speedup over $n^k$, which is presented by Sinkhorn, but is limited to relatively small $(b,k)$ values.

However, as was shown for Sinkhorn, when the cost graph is sparse, the $U$-statistic can be reformulated in terms of matrix-matrix or matrix-vector operations, leading to polynomial complexity in $(b,k)$.
We focus on a circle cost, and outline the steps to derive similar results for tree cost graphs.
When $\cG_c$ is circular, i.e., $c(\bxk)=\sum_{i=1}^k\tilde{c}(x_i,x_{i+1})$, we identify $\hat{E}_\sU$ as the trace of some matrix product, which corresponds to a loop along the circle:

\begin{proposition}\label{prop:circle_cost_loss}
Let $\cG_c$ be a circle cost graph, $\Vec{f}_i = [f_{\theta_i}(x_{i1}),\dots,f_{\theta_i}(x_{in})]\in\RR^n$ be the vector of dual potential outputs, $\rC^{i}[j,l]=\tilde{c}(x_{ij},x_{i+1,l})$ be the pairwise cost matrix for $i=1,\dots, k$ with $k+1=1$ and let $\rL_i \coloneqq \exp\left(\frac{\frac{1}{2}(\Vec{f}_i\bigoplus\Vec{f}_{i+1}) - \rC_{i}}{\epsilon}\right)\in\RR^{n\times n}$ where $\Vec{f}_i\bigoplus\Vec{f}_{i+1}$ is the direct sum of the vectors, resulting in an $n\times n$ matrix.
Then
\begin{equation}\label{eq:circle_cost_loss}
     \hat{E}_\sU =  \frac{1}{n^k}\Tr\left(\prod_{i=1}^k\rL_i\right)+\epsilon,
\end{equation}
and the complexity of calculating \eqref{eq:circle_cost_loss} is $O(kn^3)$.
Furthermore, the neural plan tensor can be represented as
\begin{equation}\label{eq:neural_plan_matmul}
    \Pi^\epsilon_{\theta^\star}(\mathbf{X}^{n,k})[i_1,\dots,i_k] = \prod_{j=1}^k \rL_j(i_j,i_{j+1}).
\end{equation}
\end{proposition}
Furthermore, we have an alternative representation of the resulting optimal transport cost, due to the representation of the neural plan:
\begin{proposition}\label{prop:unreg_ot}
     Let $(\rC^{i})_{i=1}^k$ be the pairwise cost matrices such that $\rC^{i}_{j,l}=c(x_{ij},x_{i+1,l})$ for a given circle cost graph $\cG_c$.
     The corresponding unregularized transport cost is given by
    $$
    \langle C, \widehat{\Pi}_{\theta^\star}^{\epsilon} \rangle = \sum_{j=1}^k \langle C_j, m\left(\widehat{\Pi}_{\theta^\star}^{\epsilon}, j, j+1\right) \rangle
    $$
    where $m\left(\widehat{\Pi}_{\theta^\star}^{\epsilon}, i_1,i_2\right)$ is the pairwise marginalization of $\widehat{\Pi}_{\theta^\star}^{\epsilon}$ between $(\mu_{i_1},\mu_{i_2})$, which is given by
    \begin{equation}\label{eq:circle_pairwise_plan}
        m\left(\widehat{\Pi}_{\theta^\star}^{\epsilon}, i_1,i_2\right) = \left(\prod_{j=i_1}^{i_2-1}\rL_{j}\right)   \odot 
  \left[ \left(\prod_{j=1}^{i_1-1}\rL_{j}\right)^\intercal\left(\prod_{j=i_2}^{k}\rL_{j}\right)^\intercal\right]
    \end{equation}
    where $\rA\odot\rB$ is the Hadamard product of $(\rA,\rB)$.
\end{proposition}
At the heart of Proposition \ref{prop:unreg_ot} is the efficient marginalization, which is the main computational bottleneck of the Sinkhorn algorithm.

Similar results can be drawn for the tree cost graph.
When $\cG_c$ is a tree, the resulting alternative form of $\hat{E}_\sU$ corresponds to a set of matrix-vector operations, which correspond to a DFS traversal along the tree that represents $\cG_c$.

\subsection{Proof of Proposition \ref{prop:circle_cost_loss}}\label{proof:efficient_circle_loss}
Recall that the NEMOT is given by the expression
$$
    \widehat{\MOT}_{c,\epsilon}(\textbf{X}^{n,k}) \coloneqq \sup_{\theta_1,\dots,\theta_k}\frac{1}{n}\sum_{j=1}^n \sum_{i=1}^k f_{\theta_i}(X_{ij})-\frac{\epsilon}{n^k}\sum_{j_1,\dots,j_k}e^{\left(\sum_{i=1}^k f_{\theta_i}(X_{j_i}) -c(X_{j_1},\dots,X_{j_k})\right)/\epsilon}+\epsilon.
$$
The first sum directly follows from the definition of the direct sum.
For the exponential term, we have
\begin{align*}
    &\exp\left(\frac{\sum_{i=1}^k f_{\theta_i}(X_{j_i}) -c(X_{j_1},\dots,X_{j_k})}{\epsilon}\right)\\ 
    &\hspace{4cm}= \exp\left(\frac{\sum_{i=1}^k f_{\theta_i}(X_{j_i}) - \sum_{i=1}^k \tilde{c}(X_{j_i}X_{j_{i+1}})}{\epsilon}\right)\\
    &\hspace{4cm}= \exp\left(\frac{\sum_{i=1}^k \frac{1}{2}\left(f_{\theta_i}(X_{j_i})+f_{\theta_{i+1}}(X_{j_{i+1}})\right) - \sum_{i=1}^k \tilde{c}(X_{j_i}X_{j_{i+1}})}{\epsilon}\right)\\
    &\hspace{4cm}= \prod_{i=1}^k \exp\left(\frac{ \frac{1}{2}\left(\Vec{f}_{\theta_i}[j_i]+\Vec{f}_{\theta_{i+1}}[j_{i+1}]\right) - \rC^i[j_i,j_{i+1}]}{\epsilon}\right)\\
    &\hspace{4cm}= \prod_{i=1}^k \rL_i[j_i,j_{i+1}].
\end{align*}
Note that by the definition of the neural plan \eqref{eq:neural_plan} we have obtained the desired representation.
Consequently, the second sum is given by
\begin{align}
    \sum_{j_1,\dots,j_k}\prod_{i=1}^k \rL_i[j_i,j_{i+1}] &= \sum_{j_1,\dots,j_k}\rL_1[j_1,j_{2}]\rL_2[j_2,j_{3}]\cdots\rL_k[j_k,j_{1}].
\end{align}
Note that each index we sum over corresponds only to a specific pair of matrices, and is equivalent to their matrix multiplication.
Thus, after taking the product of the set of matrices, we remain with the sum
$$
\sum_{j_1,\dots,j_k}\left(\prod_{i=1}^k \rL_i\right)[j_1,j_1] = \Tr\left( \prod_{i=1}^k \rL_i \right), 
$$
which completes the proof. $\hfill\square$

\subsection{Proof of Proposition \ref{prop:unreg_ot}}\label{app:unreg_ot_proof}
Recall that the optimal neural plan can be represented as the multiplication of exponential matrices, i.e., 
$$
\widehat{\Pi}_{\theta^\star}^{\epsilon}(i_1,i_2,\dots,i_k) = \prod_{j=1}^k \rL_j[i_j,i_{j+1}].
$$
Let $C\in\RR^{n^k}$ be the explicit cost tensor and $(\rC^{i})_{i=1}^k\subset \RR^{n\times n}$ be the corresponding pairwise cost matrices.
We begin with the representation of the unregularized cost. We have the following: 
\begin{align*}
    \langle \rC, \widehat{\Pi}_{\theta^\star}^{\epsilon} \rangle &= \sum_{i_1,i_2,\dots,i_k} C[i_1,i_2,\dots,i_k]\widehat{\Pi}_{\theta^\star}^{\epsilon}[i_1,i_2,\dots,i_k]\\
    &= \sum_{i_1,i_2,\dots,i_k} \sum_{j=1}^k \rC_j[i_j,i_{j+1}]\widehat{\Pi}_{\theta^\star}^{\epsilon}[i_1,i_2,\dots,i_k]\\
    &= \sum_{i_1,i_2,\dots,i_k} \sum_{j=1}^k \rC_j[i_j,i_{j+1}]\left(\prod_{\ell=1}^k\rL_\ell[i_\ell,i_{\ell+1}]\right)\\
    &= \sum_{j=1}^k \sum_{i_j,i_{j+1}}\rC_j[i_j,i_{j+1}]\left(\sum_{(i_1,\dots,i_k)\setminus(i_j,i_{j+1})} \prod_{\ell=1}^k\rL_\ell[i_\ell,i_{\ell+1}]\right)\\
    &= \sum_{j=1}^k \sum_{i_j,i_{j+1}}\rC_j[i_j,i_{j+1}]m\left(\widehat{\Pi}_{\theta^\star}^{\epsilon}, j, j+1\right)[i_j,i_{j+1}]\\
    &= \sum_{j=1}^k \langle \rC_j, m\left(\widehat{\Pi}_{\theta^\star}^{\epsilon}, j, j+1\right) \rangle.
\end{align*}
We now move on to prove the circle cost marginalization formula.
Let $0 \leq  u,v \leq k$ with $u\neq v$. The marginalization follows from noticing that the considered summations over the given product can be represented with products of the matrices $\rL_1,\dots,\rL_k$. For a given entry of the marginalized coupling, we have:
\begin{align*}
    m\left(\widehat{\Pi}_{\theta^\star}^{\epsilon}, u, v\right)(i_u,i_v)  = \sum_{(i_1,\dots,i_k)\setminus[i_u,i_{v}]} \left(\prod_{\ell=1}^k\rL_\ell[i_\ell,i_{\ell+1}]\right).
\end{align*}
We divide the sum into its consecutive parts and define the matrices
\begin{align*}
    \rA_{<u} \coloneqq \prod_{j=1}^{u-1} \rL_j,\qquad
    \rA_{u<v} \coloneqq \prod_{j=u+1}^{v-1} \rL_j,\qquad
    \rA_{>v} \coloneqq \prod_{j=v+1}^{k} \rL_j.
\end{align*}
We obtain the following product:
\begin{align*}
    m\left(\widehat{\Pi}_{\theta^\star}^{\epsilon}, u, v\right)[i_u,i_v] &= \sum_{i_1}\rA_{<u}[i_1,i_u]\rA_{u<v}[i_u,i_v]\rA_{>v}[i_v,i_1]\\
    &= \rA_{u<v}(i_u,i_v) \sum_{i_1}\rA_{<u}^\intercal[i_u,i_1]\rA_{>v}^\intercal[i_1,i_v]\\
    &= \rA_{u<v}[i_u,i_v] \left(\rA_{<u}^\intercal\rA_{>v}^\intercal\right)[i_u,i_v],
\end{align*}
which completes the proof. $\hfill\square$


\begin{figure}[!t]
    \centering
    \includegraphics[trim={50pt 120pt 30pt 120pt}, clip,width=0.8\linewidth]{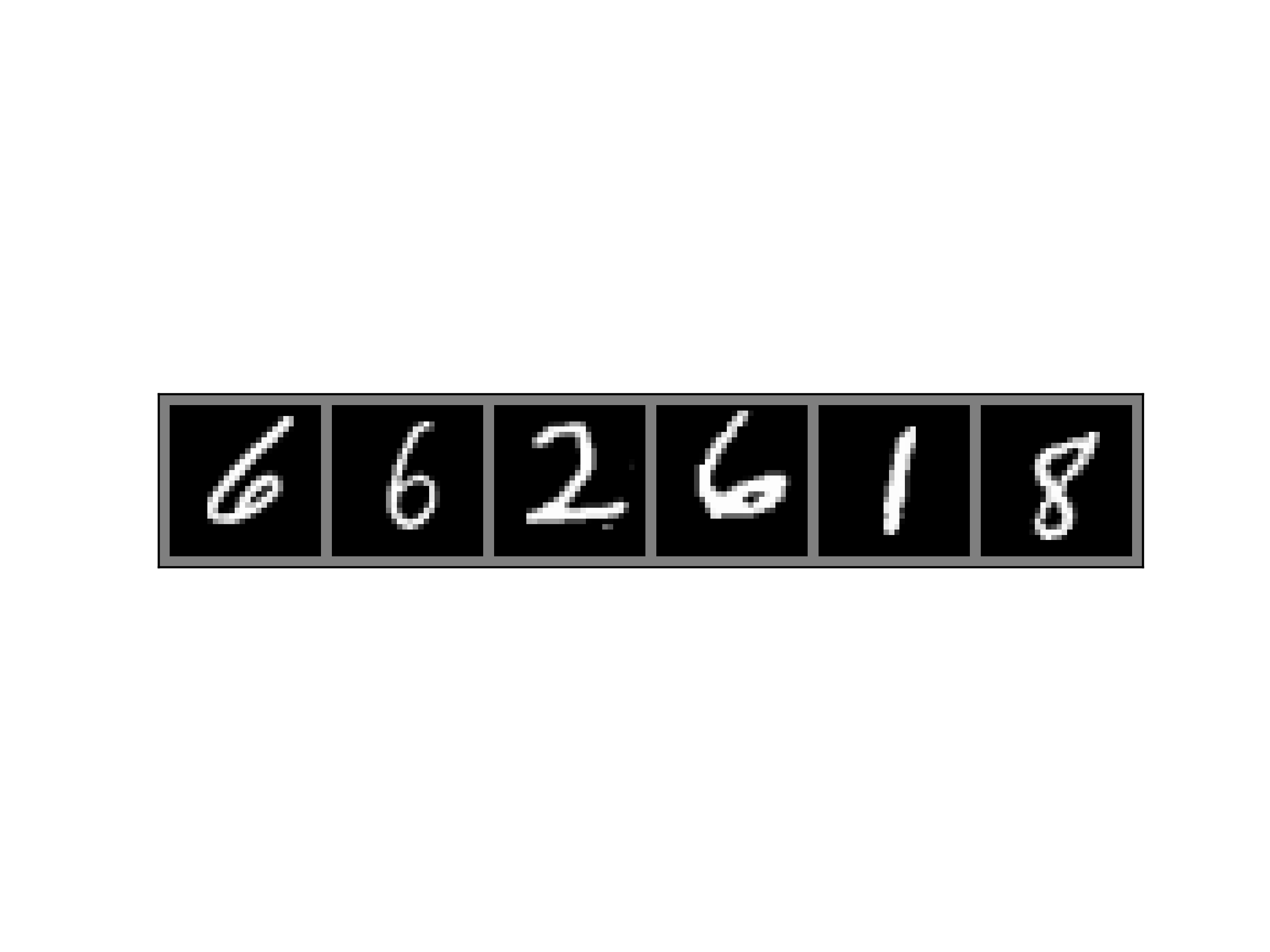}
    \caption{MNIST dataset, consisting of $28\times28$ gray-pixel images with $10$ labels.}
    \label{fig:mnist_images}
\end{figure}

\begin{figure}
    \centering
    \includegraphics[trim ={60pt 40pt 20pt 60pt}, clip,width=0.9\linewidth]{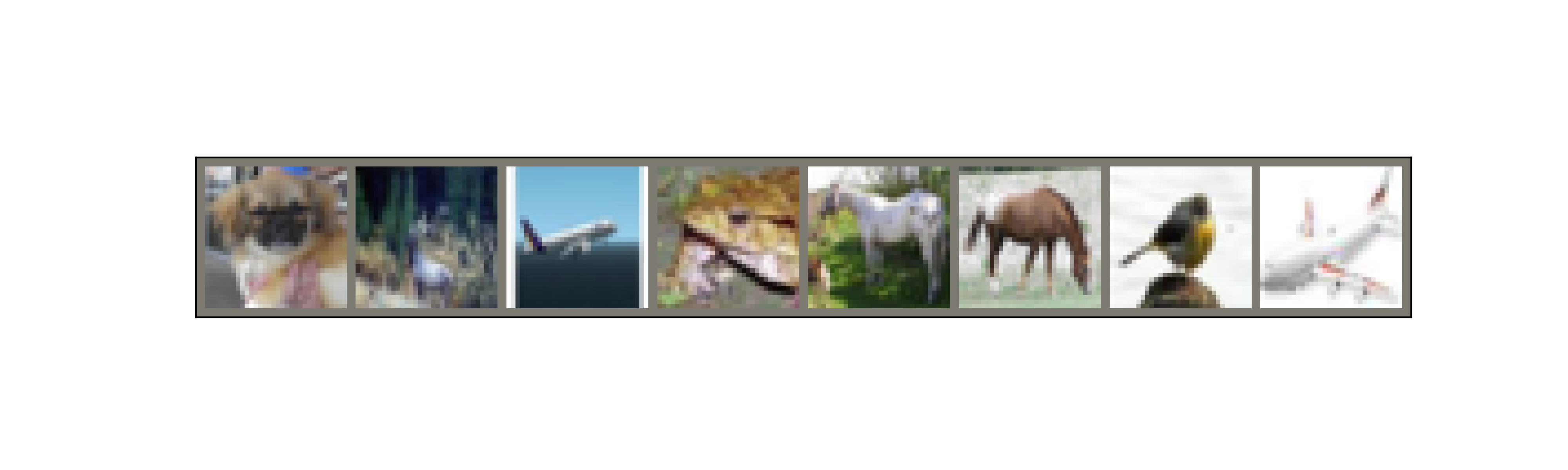}
    \caption{CIFAR10 images.}
    \label{fig:enter-label}
\end{figure}

\end{document}